\documentclass[journal, 10pt]{IEEEtran}
\IEEEoverridecommandlockouts
\usepackage{cite}
\usepackage{amsmath,amssymb,amsfonts}
\usepackage{algorithmic}
\usepackage{graphicx}
\usepackage{textcomp}
\usepackage{xcolor}
\usepackage{soul}
\usepackage{url}
\usepackage{wrapfig}
\usepackage{balance}
\usepackage{scalerel}

\usepackage{colortbl}
\definecolor{Gray}{gray}{0.925}

\usepackage[most]{tcolorbox}
\definecolor{shadecolor}{rgb}{0.99,0.99,0.99}
\definecolor{framecolor}{rgb}{0.25,0.25,0.25}
\newtcolorbox{leftlinebox}{
    enhanced,
    colback=shadecolor,
    colframe=white, 
    borderline west={1mm}{0mm}{framecolor},
    boxrule=0pt,
    width=\linewidth,
    arc=1.25mm,
    boxsep=2pt,   
    left=4pt,
    right=3pt,
    top=2pt,
    bottom=2pt,
    enlarge top by=-4.75pt,
    enlarge bottom by=-4.75pt
}

\usepackage{fontawesome}

\def\BibTeX{{\rm B\kern-.05em{\sc i\kern-.025em b}\kern-.08em
    T\kern-.1667em\lower.7ex\hbox{E}\kern-.125emX}}
    
\usepackage{tikz}

\usepackage{adjustbox}
\usepackage{array}
\usepackage{booktabs}
\usepackage{multirow}
\newcolumntype{R}[2]{%
    >{\adjustbox{angle=#1,lap=\width-(#2)}\bgroup}%
    l%
    <{\egroup}%
}
\newcommand*\rot{\multicolumn{1}{R{45}{1em}}}%
\usepackage{bbding}
\usepackage{pifont}
\newcommand{\cmark}{\color{green} \ding{51}}%
\newcommand{\xmark}{\color{red} \ding{55}}%

\usepackage[font=small]{caption}

\usepackage{enumitem}
\setlist{leftmargin=5mm}

\definecolor{spc}{rgb}{0.935, 0.58, 0.0}
\definecolor{pct}{rgb}{0.31115724721261057, 0.6082891195693964, 0.7968935024990389}

\begin{document}

\title{
Exploring Deep-to-Shallow Transformable Neural Networks for Intelligent Embedded Systems
}

\author{Xiangzhong~Luo$^*$
        and Weichen~Liu, \textit{Member, IEEE}
        \thanks{
            $^*$The corresponding author is Xiangzhong Luo.

            This work is supported by the Basic Research Program of Jiangsu (BK20251309), the Startup Project of Southeast University (4009002505), and the Big Data Computing Center of Southeast University. This work is partially supported by the Ministry of Education, Singapore, under its Academic Research Fund Tier 2 (MOE-T2EP20224-0006) and Tier 1 (RG94/23).

            Xiangzhong Luo is currently with the School of Computer Science and Engineering, Southeast University (SEU), China. (Email: xiangzhong.luo@seu.edu.cn)

            Weichen Liu is currently with the College of Computing and Data Science, Nanyang Technological University (NTU), Singapore. (Email: liu@ntu.edu.sg)
        }
}

\maketitle

\begin{abstract}
Thanks to the evolving network depth, convolutional neural networks (CNNs) have achieved remarkable success across various embedded scenarios, paving the way for ubiquitous embedded intelligence. Despite its promise, the evolving network depth comes at the cost of degraded hardware efficiency. In contrast to deep networks, shallow networks can deliver superior hardware efficiency but often suffer from inferior accuracy. To address this dilemma, we propose Double-Win NAS, a novel deep-to-shallow transformable neural architecture search (NAS) paradigm tailored for resource-constrained intelligent embedded systems. Specifically, Double-Win NAS strives to automatically explore deep networks to first win strong accuracy, which are then equivalently transformed into their shallow counterparts to further win strong hardware efficiency. In addition to search, we also propose two enhanced training techniques, including hybrid transformable training towards better training accuracy and arbitrary-resolution elastic training towards enabling natural network elasticity across arbitrary input resolutions. Extensive experimental results on two popular intelligent embedded systems (i.e., NVIDIA Jetson AGX Xavier and NVIDIA Jetson Nano) and two representative large-scale datasets (i.e., ImageNet and ImageNet-100) clearly demonstrate the superiority of Double-Win NAS over previous state-of-the-art NAS approaches.
\end{abstract}

\section{Introduction}
\label{sec:introduction}

Convolutional neural networks (CNNs) have empowered a myriad of embedded scenarios towards ubiquitous embedded intelligence \cite{liu2022bringing}, such as on-device object detection/tracking \cite{neseem2021adacon} and immersive AR/VR \cite{li2022rt}. In the era of deep learning, \textit{deeper + wider = better} has been empirically deemed as the rule of thumb to design novel networks with promising accuracy, as shown in previous representative networks \cite{sandler2018mobilenetv2, ma2018shufflenet, he2016deep}. Some recent studies \cite{nguyen2020wide, xue2022go} also demonstrate that increasing the network depth or width, when properly engineered, can largely improve the attainable accuracy. However, this empirical rule, despite its promise, requires substantial engineering efforts from human experts to manually explore the optimal network structure \cite{luo2024efficient, benmeziane2021comprehensive}. To overcome such limitations, recent network design practices have shifted from \textit{manual} to \textit{automated}, also known as neural architecture search (NAS), which is dedicated to automatically exploring novel network structures. The searched networks have been shown to push forward state-of-the-art accuracy on target task \cite{liu2018darts, tan2019mnasnet, howard2019searching, chu2020moga, cai2019once}. Among them, differentiable NAS \cite{liu2018darts} has dominated recent success in the field of NAS compared with other representative NAS variants (e.g., reinforcement learning based NAS and evolutionary algorithm based NAS), thanks to its reliable search performance and strong search efficiency \cite{benmeziane2021comprehensive}.

The dominant success of differentiable NAS \cite{liu2018darts} has subsequently sparked a plethora of \textbf{H}ard\textbf{W}are-aware \textbf{D}ifferentiable \textbf{NAS} (HW-DNAS) methods to search for hardware-friendly network solutions \cite{li2020edd, zhang2021dian, luo2022lightnas, cai2018proxylessnas, wu2019fbnet, stamoulis2019single, hu2020tfnas, fang2020densely, luo2022you, li2022physics}. Specifically, previous state-of-the-art (SOTA) HW-DNAS methods \cite{li2020edd, zhang2021dian, luo2022lightnas, cai2018proxylessnas, wu2019fbnet, stamoulis2019single, hu2020tfnas, fang2020densely, luo2022you, li2022physics} typically feature sufficient network depth to explore deep networks in order to maintain decent accuracy \cite{benmeziane2021comprehensive}. However, the resulting deep networks, despite their strong accuracy, exhibit degraded hardware efficiency \cite{he2015convolutional}. The rationale here is that the deployed network must follow layer-wise execution patterns on modern hardware processors and its lowest achievable latency is $D/F$, where $D$ is the network depth and $F$ is the processor frequency \cite{goyal2022non}. To remedy this issue, we may simply re-engineer previous SOTA HW-DNAS methods to explore shallow networks instead. However, the resulting shallow networks, despite their superior hardware efficiency, suffer from inferior accuracy since the attainable accuracy relies on sufficient network depth \cite{he2016deep}. This dilemma reveals that previous SOTA HW-DNAS methods can only explore deep or shallow networks to win either accuracy or hardware efficiency and cannot deliver an aggressive win-win in terms of both. To tackle this dilemma, we turn to the following counterintuitive question:
\begin{tcolorbox}[colback=shadecolor, boxrule=0.4mm,left=2pt,right=2pt,top=2pt,bottom=2pt,
                grow to left by=-1pt,
                grow to right by=-1pt,
                enlarge top by=-3pt,
                enlarge bottom by=-3pt,
                title=\textit{\faHandORight\,\,\,\,Question I: Transformable Network Search},
                fonttitle=\small]
\textit{How can we search for deep networks to first win accuracy, which can be further equivalently transformed into their shallow counterparts to win hardware efficiency and finally enable an aggressive accuracy-efficiency win-win?}
\end{tcolorbox}

\begin{figure*}[t]
    \begin{center}
        \includegraphics[width=1.0\linewidth]{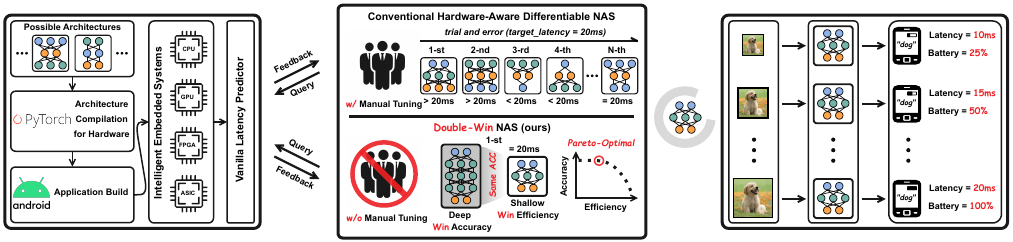}
    \end{center}
    \vspace{-4pt}
    \caption{An intuitive overview of Double-Win NAS (DW-NAS) that focuses on exploring and training deep-to-shallow transformable networks.}
    \vspace{-6pt}
    \label{fig:overview}
\end{figure*}

Furthermore, previous SOTA HW-DNAS methods \cite{li2020edd, zhang2021dian, luo2022lightnas, cai2018proxylessnas, wu2019fbnet, stamoulis2019single, hu2020tfnas, fang2020densely, luo2022you, li2022physics} focus on training the searched networks using standard training recipes. Nonetheless, the resulting well-trained networks can only accommodate fixed computational budgets and cannot naturally achieve elastic accuracy-efficiency trade-offs during runtime executions \cite{yu2018slimmable, yu2019universally, wang2018skipnet, yang2022once, wang2020resolution, ghebriout2024harmonic}. Among them, Harmonic-NAS \cite{ghebriout2024harmonic} considers elastic accuracy-efficiency trade-offs during the search phase to enhance the search efficiency, which, however, cannot generalize to inference on target hardware. We note that the natural network elasticity is of paramount importance for real-world embedded scenarios, such as battery-powered mobile phones, where the available computational resources may dramatically change over time \cite{luo2024efficient}. To alleviate the above issue, the first-in-mind solution is to train multiple stand-alone networks under different computational budgets and then specialize them on target hardware for elastic inference at runtime \cite{yu2018slimmable, yu2019universally, wang2018skipnet, yang2022once, wang2020resolution}. However, this may suffer from non-trivial training overheads and considerable on-device storage requirements \cite{yu2018slimmable, yu2019universally, wang2018skipnet, yang2022once, wang2020resolution}. To overcome such limitations, \cite{yu2018slimmable, yu2019universally, wang2018skipnet, yang2022once, wang2020resolution} have been recently proposed, which focus on enabling natural network elasticity across diverse network widths \cite{yu2018slimmable, yu2019universally}, depths \cite{wang2018skipnet, yang2022once}, and input resolutions \cite{wang2020resolution}. In contrast to \cite{yu2018slimmable, yu2019universally, wang2018skipnet, yang2022once}, RS-Net \cite{wang2020resolution} does not introduce additional costs during runtime resolution switching. However, RS-Net only supports a limited number of input resolutions and cannot aggressively generalize to arbitrary input resolutions. To tackle this dilemma, we turn to the following counterintuitive question:

\begin{tcolorbox}[colback=shadecolor, boxrule=0.4mm,left=2pt,right=2pt,top=2pt,bottom=2pt,
                grow to left by=-1pt,
                grow to right by=-1pt,
                enlarge top by=-3pt,
                enlarge bottom by=-3pt,
                title=\textit{\faHandORight\,\,\,\,Question II: Elastic Network Training},
                fonttitle=\small]
\textit{How can we train the searched transformable network to unlock natural network elasticity towards elastic accuracy-efficiency trade-offs across arbitrary input resolutions?}
\end{tcolorbox}

To investigate the above questions, we establish the first-of-its-kind deep-to-shallow transformable HW-DNAS paradigm, namely Double-Win NAS (DW-NAS), pioneering to draw insights from deep-to-shallow transformable networks to marry the best of both deep and shallow networks towards an aggressive accuracy-efficiency win-win. As shown in Fig.~\ref{fig:overview} (\textit{left}), DW-NAS strives to explore deep networks to first win strong accuracy, which then can be equivalently transformed into their shallow counterparts to win considerable hardware efficiency, and more importantly, without accuracy loss. Furthermore, as shown in Fig.~\ref{fig:overview} (\textit{right}), we also introduce a simple yet effective elastic training paradigm to unlock natural network elasticity, which can enable elastic accuracy-efficiency trade-offs during runtime executions across arbitrary input resolutions. Finally, we summarize our novel contributions as follows:
\begin{itemize}
    \renewcommand\labelitemi{\faThumbsOUp}
    \item {
    \textbf{Fundamentals.}
    To the best of our knowledge, the proposed DW-NAS is the first deep-to-shallow transformable NAS paradigm, which opens up a fresh NAS perspective and also provides a \textit{\textbf{holistic}} solution for \textit{\textbf{designing}} and \textit{\textbf{training}} deep-to-shallow transformable networks towards \textit{\textbf{efficient}} and \textit{\textbf{elastic}} embedded intelligence.
    }
    \renewcommand\labelitemi{\faHandORight}
    \item {
    \textbf{Framework.}
    The proposed DW-NAS features a novel hybrid transformable search space to achieve deep-to-shallow transformable search, which is integrated with an efficient vanilla latency predictor and an effective sandwich-inspired differentiable search algorithm to explore the optimal transformable network around the specified latency constraint. Furthermore, we also draw insights from the searched transformable network and introduce two enhanced training techniques, including hybrid transformable training towards better training accuracy and arbitrary-resolution\footnote{In this work, the definition of arbitrary resolution refers to any resolution ranging from $96\times96$ to $224\times224$ with an interval of $8\times8$, which aligns with hardware tiling and stride-2 down-sampling constraints.} elastic training towards runtime resolution switching across arbitrary input resolutions. Finally, the searched transformable network can be equivalently transformed into its shallow counterpart to achieve aggressive on-device efficiency improvement without accuracy loss.
    }
    \renewcommand\labelitemi{\faHandORight}
    \item {
    \textbf{Evaluations.} 
    Extensive experiments and ablation studies are conducted to evaluate the proposed DW-NAS on two popular NVIDIA Jetson intelligent embedded systems (i.e., Xavier and Nano) and two representative large-scale datasets (i.e., ImageNet and ImageNet-100), which clearly demonstrate the superiority of DW-NAS over previous SOTA NAS approaches. For example, the searched DW-Net-Xavier-20ms can achieve $+$0.8\% higher accuracy on ImageNet than FBNet-C \cite{wu2019fbnet}, while at the same time maintaining $\times$1.3 inference speedup on Xavier.
    }
\end{itemize}    


\section{Background}
\label{sec:background}

In this section, we first introduce the preliminaries on differentiable NAS \cite{liu2018darts, wu2019fbnet}, after which we discuss the observations and motivations behind the proposed approach.

\subsection{Preliminaries on Differentiable NAS}
\label{sec:preliminaries-on-differentiable-nas}

Let $\mathcal{O} = \{o_n\}_{n=1}^N$ represent the operator space with $N$ pre-defined operator candidates. Following DARTS \cite{liu2018darts}, an over-parameterized network dubbed supernet is first initialized. Note that the supernet is equivalent to the search space in the context of NAS since the supernet encodes all the possible architecture candidates. As defined in \cite{wu2019fbnet}, the supernet consists of $L$ searchable layers, each of which contains $N$ operator candidates $\{o_n\}_{n=1}^N$. Furthermore, to relax the discrete search space to be continuous, a set of architecture parameters $\alpha \in \mathbb{R}^{L \times N}$ are also assigned to the operators in the supernet, as discussed in \cite{liu2018darts, wu2019fbnet}. Finally, we can mathematically formulate the $l$-th searchable layer of the supernet as follows:
\begin{equation}
    x_{l+1} = \sum_{n=1}^{N} p_l^n \cdot o_n(x_l), \,\, \mathrm{s.t.}, \,\,  p_l^n = \frac{\exp(\alpha_l^n)}{\sum_{n'=1}^N \exp(\alpha_l^{n'})}
    \label{eq:darts-softmax-relaxation}
\end{equation}
where $x_l$ is the input of the $l$-th searchable layer. Thanks to the above continuous relaxation, both architecture parameters $\alpha$ and network weights $w$ can be optimized using gradient descent. In light of this, DARTS further introduces the following bi-level differentiable optimization strategy to alternate the differentiable optimization process of $\alpha$ and $w$:
\begin{equation}
    \begin{aligned}
    &\mathop{\mathrm{minimize}}_{\alpha} \, \mathcal{L}_{valid}(w^*(\alpha), \alpha)\\
    &\,\,\mathrm{s.t.}, \,\, w^*(\alpha)=\mathop{\mathrm{arg\,min}}_{w} \, \mathcal{L}_{train}(w, \alpha)
    \end{aligned}
    \label{eq:darts-optimization}
\end{equation}
where $\mathcal{L}_{train}(\cdot)$ and $\mathcal{L}_{valid}(\cdot)$ are the training and validation loss functions. Once the above bi-level differentiable optimization converges, we can discretize the searched optimal architecture according to the learned architecture parameters $\alpha$. The discretization process reserves the strongest operator in each searchable layer and eliminates the remaining operators from the supernet, in which the operator strength of $o_n$ in the $l$-th searchable layer is defined as $\exp(\alpha_{l}^{n})/\sum_{n'=1}^{N} \exp(\alpha_{l}^{n'})$ \cite{liu2018darts, wu2019fbnet}. The interested readers may refer to DARTS \cite{liu2018darts} or FBNet \cite{wu2019fbnet} for more details about differentiable NAS.

\begin{figure}[t]
    \begin{center}
        \includegraphics[width=1.0\columnwidth]{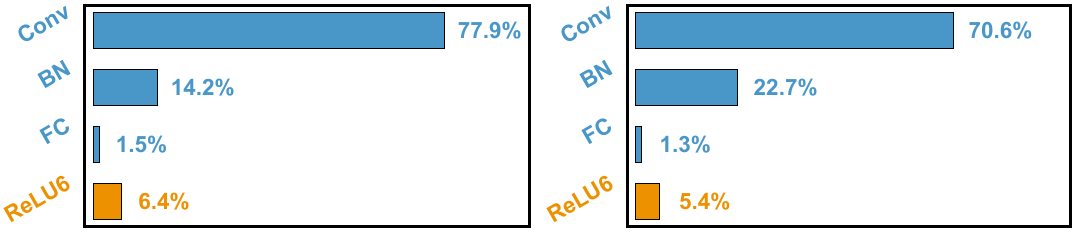}
    \end{center}
    \vspace{-4pt}
    \caption{The layer-wise latency profiling analysis of MobileNetV2 on Xavier (\textit{left}) and Nano (\textit{right}), where only ReLU6 is non-linear. The reported percentage corresponds to the latency percentage of different layers, compared with the total latency of MobileNetV2.}
    \vspace{-6pt}
    \label{fig:mobilenetv2-profiling-results}
\end{figure}

\subsection{Observations and Motivations}
\label{sec:observations-and-motivations}

\textbf{Observation~I.}
\textit{Multiple consecutive linear layers can be equivalently transformed into one single linear layer without changing the output.} Without loss of generality, we consider the following two consecutive linear layers:
\begin{equation}
    Y = W_1 X + B_1 \,\,\, \mathrm{and} \,\,\, Z = W_2 Y + B_2
    \label{eq:two-linear-layers}
\end{equation}
where $X$ is the input. Besides, $W_1$, $W_2$, $B_1$, and $B_2$ are the weight and bias parameters of the above two consecutive linear layers. Taken together, we can re-formulate Eq~(\ref{eq:two-linear-layers}) as follows:
\begin{equation}
    Z = W_2  (W_1 X + B_1) + B_2= (W_1 W_2) X + (W_2 B_1 + B_2)
    \label{eq:one-linear-layer}
\end{equation}
This indicates that the above two consecutive linear layers can be equivalently transformed into one single linear layer with $W^* = W_1 W_2$ and $B^* = W_2 B_1 + B_2$, which can also maintain the same output $Z$. Note that Eq~(\ref{eq:two-linear-layers})-(\ref{eq:one-linear-layer}) can be generalized to all common linear layers, such as convolutional (\texttt{Conv}), batch normalization (\texttt{BN}), and fully-connected (\texttt{FC}) layers. For example, one \texttt{Conv} layer and its subsequent \texttt{Conv} or \texttt{BN} layer can be transformed into one single \texttt{Conv} layer \cite{guo2020expandnets, luo2024pearls}.

\textbf{Observation~II.} 
\textit{The linear layers largely dominate the runtime latency on target hardware.}
To this end, we profile MobileNetV2 \cite{sandler2018mobilenetv2} on Xavier and Nano with an input batch size of 8, where the inter-layer communication overheads and data movements are ignored for simplicity \cite{luo2024pearls}. In practice, we first replace all its ReLU6 layers with computation-free identity mapping, after which we can measure the latency of the resulting MobileNetV2-w/o-ReLU6 on Xavier and Nano. Based on this, we can easily calculate the total latency corresponding to all ReLU6 layers. Similar to ReLU6, we can also calculate the total latency corresponding to all \texttt{Conv} layers, all \texttt{BN} layers, and all \texttt{FC} layers, respectively. The profiling results are shown in Fig.~\ref{fig:mobilenetv2-profiling-results}, which clearly demonstrates that the linear layers, especially \texttt{Conv} layers, account for over 93\% of the total runtime latency on both Xavier and Nano. This also indicates that the findings in Observation~I have the potential to deliver considerable on-device speedups.

\begin{figure}[t]
    \begin{center}
        \includegraphics[width=1.0\columnwidth]{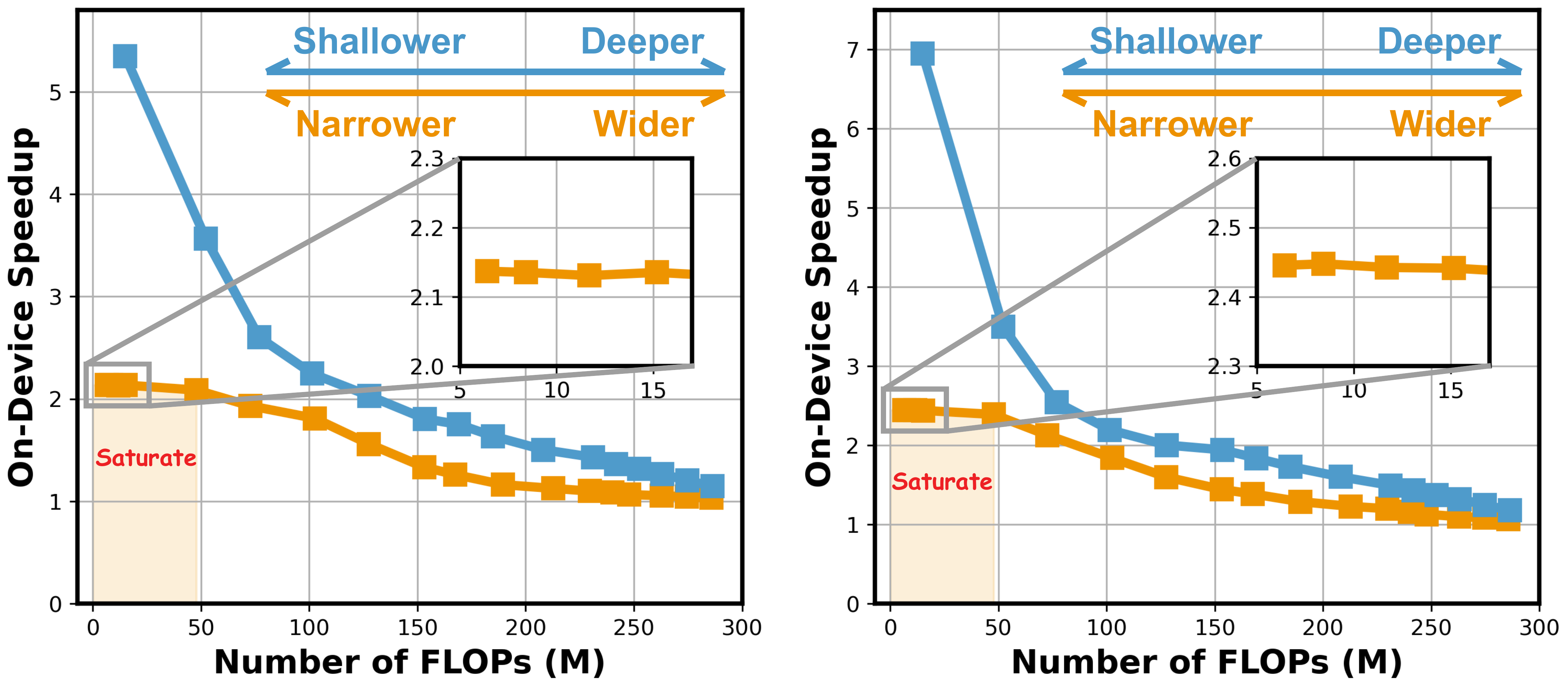}
    \end{center}
    \vspace{-4pt}
    \caption{Comparisons of shallow and narrow networks on Xavier (\textit{left}) and Nano (\textit{right}). Note that the experimental settings are similar to \cite{luo2024pearls}. The blue and orange nodes correspond to the speedups of shallow and narrow networks derived from layer-wise pruning and channel-wise pruning, compared with the original MobileNetV2.}
    \vspace{-6pt}
    \label{fig:shallow-vs-narrow}
\end{figure}

\textbf{Observation~III.}
\textit{Hardware-friendly networks should be shallow rather than narrow.} 
Taking MobileNetV2 as an example, we first employ layer-wise and channel-wise pruning to gradually reduce the network depth and width to explore shallow and narrow networks with less floating-point operations (FLOPs). Similar to the settings in \cite{luo2024pearls}, the resulting shallow and narrow networks are then deployed on Xavier and Nano to measure their on-device latency. Finally, we can calculate the on-device speedups of these shallow and narrow networks, compared with the original MobileNetV2. As illustrated in Fig.~\ref{fig:shallow-vs-narrow}, under similar FLOPs, shallow networks exhibit better speedups than their narrow counterparts on both Xavier and Nano. Also, we find that, for narrow networks, their speedups saturate at large pruning ratios, whereas shallow networks consistently deliver enhanced speedups. The above empirical findings clearly reveal the merits of shallow networks in pursuit of superior hardware efficiency \cite{he2015convolutional, luo2024pearls}.

\begin{figure}[t]
    \begin{center}
        \includegraphics[width=1.0\columnwidth]{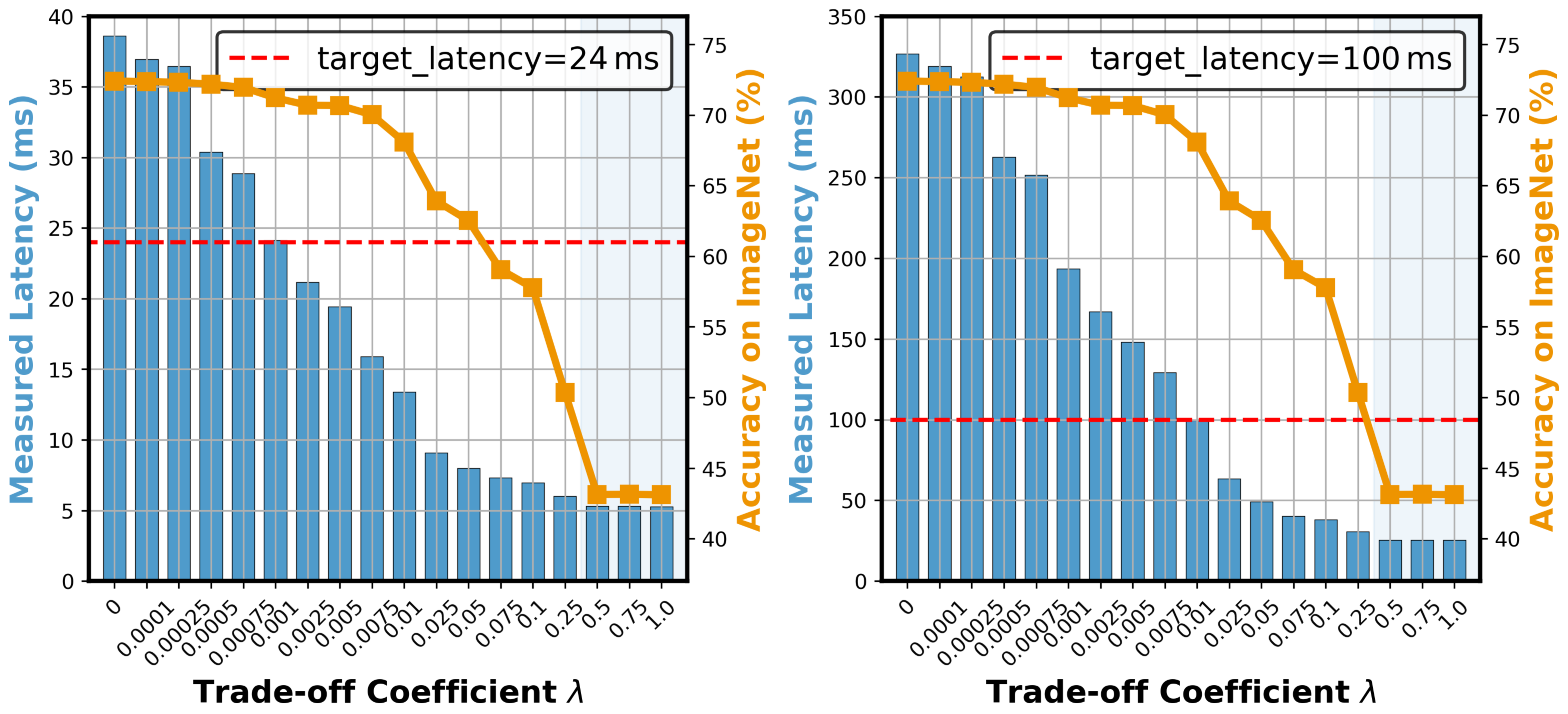}
    \end{center}
    \vspace{-4pt}
    \caption{Architecture search results under $\lambda \in [0, 1]$ on Xavier (\textit{left}) and Nano (\textit{right}), in which the latency is measured with an input batch size of 8 and the accuracy is trained on ImageNet for 50 epochs.}
    \vspace{-6pt}
    \label{fig:lambda-tradeoff}
\end{figure}

\textbf{Observation~IV.}
\textit{Previous SOTA HW-DNAS methods must repeat multiple search runs to navigate the required architecture.} As discussed in \cite{benmeziane2021comprehensive}, previous SOTA HW-DNAS methods typically introduce the following multi-objective optimization to explore hardware-efficient architecture solutions:
\begin{equation}
    \mathop{\mathrm{minimize}}_{\alpha} \,\, \mathcal{L}_{valid}(w^*(\alpha), \alpha) + \lambda \cdot LAT(\alpha)
    \label{eq:proxylessnas-objective}
\end{equation}
where $LAT(\alpha)$ is the latency of the architecture encoded by $\alpha$. Besides, $\lambda \geq 0$ is a constant to control the trade-off magnitude between accuracy and latency. In practice, the above multi-objective optimization can derive hardware-efficient architecture solutions with superior accuracy-efficiency trade-offs \cite{benmeziane2021comprehensive}, but only when $\lambda$ is properly engineered. To show such limitations, we leverage the search algorithm of FBNet \cite{wu2019fbnet} to perform a series of search experiments under $\lambda \in [0, 1]$, in which the specified latency constraint is set to 24\,ms on Xavier and 100\,ms on Nano, respectively. As shown in Fig.~\ref{fig:lambda-tradeoff}, $\lambda$ can strike trade-offs between accuracy and latency, which, however, is quite sensitive. As a result, to derive the required architecture around the specified latency, we have to repeat multiple search runs to manually tune $\lambda$ through trial and error, which is inflexible and incurs prohibitive search cost.

\begin{figure}[t]
    \begin{center}
        \includegraphics[width=1.0\columnwidth]{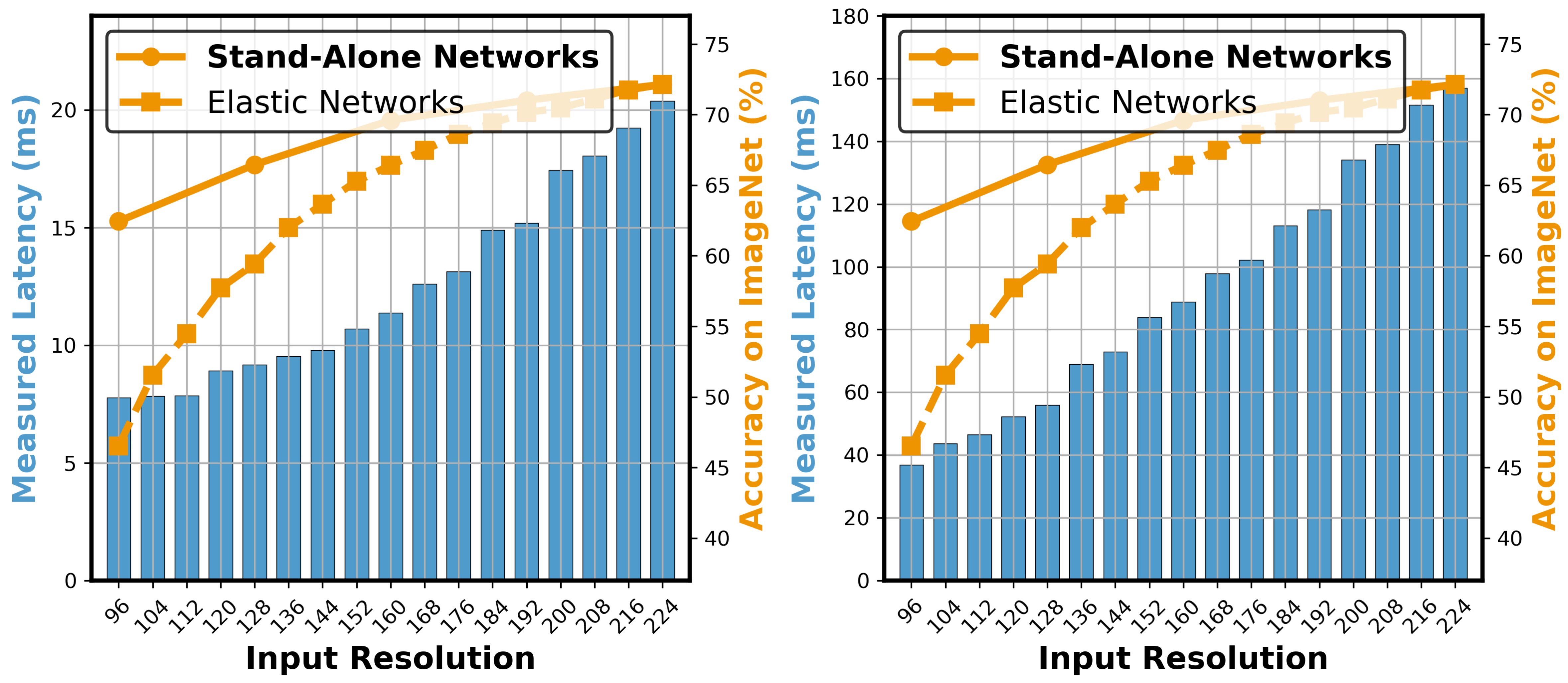}
    \end{center}
    \vspace{-4pt}
    \caption{Comparisons of the accuracy of MobileNetV2 on ImageNet under elastic and stand-alone settings across diverse input resolutions, where the latency is measured on Xavier (\textit{left}) and Nano (\textit{right}).}
    \vspace{-6pt}
    \label{fig:mobilenetv2-elastic-imagenet-motivations}
\end{figure}

\textbf{Observation~V.}
\textit{Pre-trained networks are not naturally elastic across unseen input resolutions.} 
Taking MobileNetV2 \cite{sandler2018mobilenetv2} as an example, we investigate its attainable accuracy on ImageNet across diverse input resolutions from $96\times96$ to $224\times224$ with an interval of $8\times8$\footnote{In this work, we do not consider other input resolutions that cannot be divided by 8, which cannot achieve realistic on-device speedups due to the intermediate down-sampling convolutional layers with the stride of 2.}, where the network weights are trained under the fixed input resolution of $224\times224$. The experimental results are illustrated in Fig.~\ref{fig:mobilenetv2-elastic-imagenet-motivations}, which shows that enabling network elasticity across diverse input resolutions can achieve considerable/consistent speedups on both Xavier and Nano. However, compared with stand-alone networks, this may suffer from non-negligible accuracy loss on ImageNet, especially under small input resolutions (e.g., $96\times96$). This demonstrates that networks trained under fixed input resolutions cannot naturally adapt to unseen input resolutions due to the train-test resolution discrepancy \cite{touvron2019fixing} and thus fail to enable elastic accuracy-efficiency trade-offs at runtime.

{\large{\faLightbulbO}}
\textbf{Motivations.}
As discussed in Observation~I/II/III, deep networks with multiple consecutive linear layers can be equivalently transformed into shallow networks, which may largely enhance the on-device efficiency without sacrificing the attainable accuracy. Despite its efficacy, this is infeasible because linear and non-linear layers typically alternate in order to maintain superior accuracy as shown in recent representative networks \cite{he2016deep, sandler2018mobilenetv2, ma2018shufflenet}. This motivates us to investigate deep-to-shallow transformable networks featuring multiple consecutive linear layers. With this in mind, we establish the first deep-to-shallow transformable HW-DNAS paradigm dubbed Double-Win NAS (DW-NAS), pioneering to explore deep-to-shallow transformable networks to marry the best of both deep and shallow networks towards an aggressive accuracy-efficiency win-win. Furthermore, in contrast to previous SOTA HW-DNAS methods \cite{cai2018proxylessnas, wu2019fbnet, stamoulis2019single} that have to repeat multiple search runs (see Observation~IV), DW-NAS strives to navigate the optimal deep-to-shallow transformable network around the specified latency constraint in one single search run, which can deliver considerable search efficiency and flexibility. Finally, as discussed in Observation~V, pre-trained networks are not naturally elastic across other unseen input resolutions. In contrast to previous SOTA HW-DNAS methods \cite{cai2018proxylessnas, wu2019fbnet, stamoulis2019single} that only support one fixed input resolution and RS-Net \cite{wang2020resolution} that only supports limited input resolutions, we focus on enabling natural network elasticity across arbitrary input resolutions.

\section{Double-Win NAS}
\label{sec:double-win-nas}

In this section, we first elaborate on the proposed Double-Win NAS (DW-NAS) framework and then discuss the relationships with previous SOTA HW-DNAS methods.

\subsection{Hybrid Transformable Search Space}
\label{sec:hybrid-transformable-search-space}

\begin{figure}[t]
    \begin{center}
        \includegraphics[width=1.0\columnwidth]{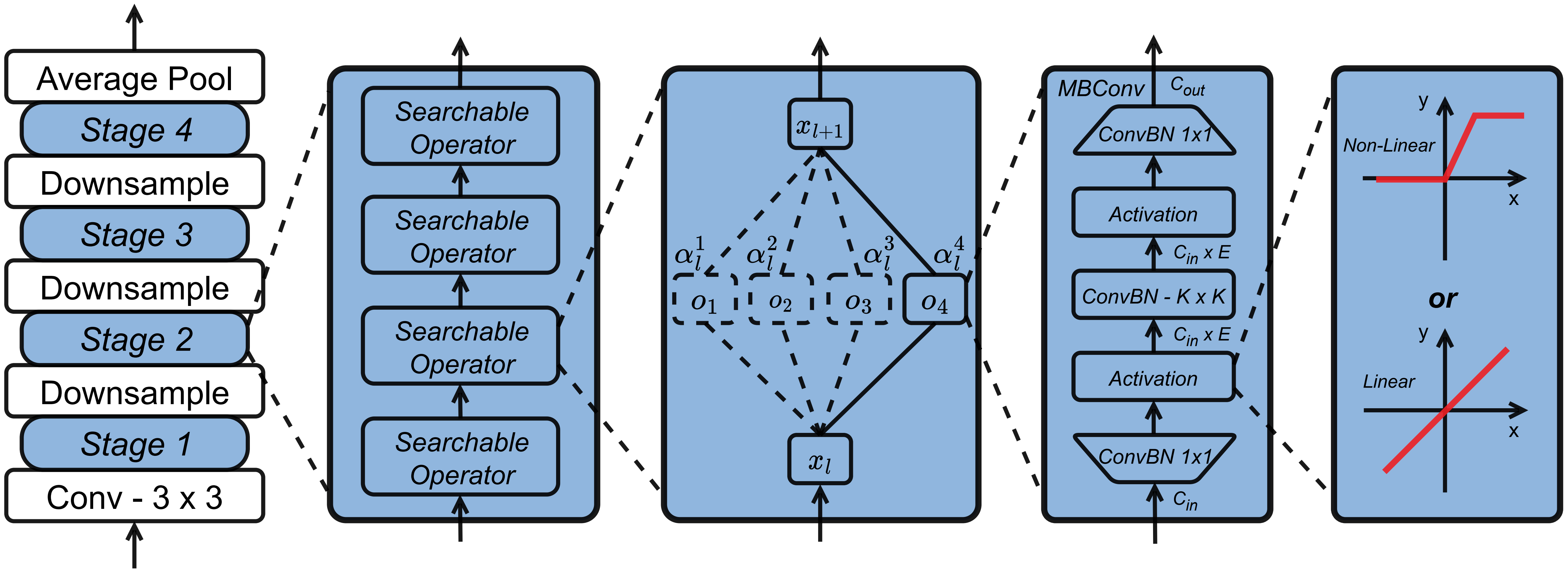}
    \end{center}
    \vspace{-4pt}
    \caption{Illustration of the hybrid transformable search space, where $K$ and $E$ denote the kernel size and the expansion ratio, respectively. Note that both activations share the same linearity or non-linearity.}
    \vspace{-6pt}
    \label{fig:search-space}
\end{figure}

Following recent well-established HW-DNAS conventions \cite{wu2019fbnet, cai2018proxylessnas, stamoulis2019single}, we construct the hybrid transformable search space $\mathcal{A}$ upon MobileNetV2. For fair comparisons with MobileNetV3-style models, we also introduce the Squeeze-and-Excitation (SE) module \cite{hu2018squeeze}, which serves as an enhancement and can be seamlessly integrated into the searched deep-to-shallow transformable networks to boost their training accuracy. As shown in Fig.~\ref{fig:search-space}, the operator space $\mathcal{O}$ consists of various linear and non-linear \texttt{MBConv} operators with diverse kernel sizes of $K \in \{3, 5, 7\}$ and expansion ratios of $E \in \{3, 6\}$. Among them, each linear \texttt{MBConv} operator consists of multiple consecutive linear layers, including three \texttt{Conv} layers, three \texttt{BN} layers, and two linear activation layers, which can be equivalently transformed into one single \texttt{Conv} layer to enable deep-to-shallow transformation without accuracy loss (see Observation~I). In this work, the implementation of \texttt{MBConv} is built upon ProxylessNAS \cite{cai2018proxylessnas}. Furthermore, we also introduce an additional non-linear ReLU6 activation layer at the end of each linear \texttt{MBConv} operator, which can largely enhance the attainable accuracy of the searched transformable network with only negligible computational overheads as discussed in Observation~II. Note that each linear \texttt{MBConv} operator featuring the last non-linear ReLU6 activation enhancement can still be equivalently transformed into one single \texttt{Conv} layer. Taken together, we have $|\mathcal{O}|=2\times3\times2=12$, and considering that the supernet has $L=21$ searchable layers \cite{cai2018proxylessnas}, we can interpret that the hybrid transformable search space consists of $|\mathcal{A}|=12^{21}\approx4.6\times10^{22}$ different transformable architecture candidates.

\subsection{Vanilla Latency Prediction}
\label{sec:vanilla-latency-prediction}

As discussed in Section~\ref{sec:hybrid-transformable-search-space}, the proposed hybrid transformable search space $\mathcal{A}$ consists of $|\mathcal{A}|\approx4.6\times10^{22}$~transformable architecture candidates, which requires non-trivial computational resources for exhaustive on-device latency measurements \cite{cai2018proxylessnas, wu2019fbnet, stamoulis2019single}. To overcome such limitations, we closely follow previous relevant practices \cite{luo2022lightnas} and introduce a simple yet effective vanilla latency predictor to reliably predict the on-device latency over the proposed hybrid transformable search space. This can effectively avoid tremendous repeated on-device latency measurements during the search process. To this end, we first encode possible transformable architecture candidates with the following sparse matrix $\overline{\alpha} \in \mathbb{R}^{L \times N}$:
\begin{equation}
    \overline{\alpha}_l^n = \mathrm{one\_hot}(\alpha) = \begin{cases}
        1, & \mathrm{if} \,\,\, \alpha_{l}^{n} = \mathop{\mathrm{arg\,max}}_n ||\alpha_{l}|| \\
        0, & \mathrm{otherwise}
    \end{cases}
    \label{eq:alpha-sparse-matrix}
\end{equation}
where $\overline{\alpha}_l^n=1$ indicates that the operator $o_n$ is reserved in the $l$-th searchable layer while the remaining operators with $\overline{\alpha}=0$ are eliminated. With the above in mind, we further introduce the proposed vanilla latency predictor $LAT(\cdot)$, which takes $\overline{\alpha}$ as input to predict the on-device latency $LAT(\overline{\alpha})$. Specifically, the vanilla latency predictor features a simple multi-layer perceptron (MLP) model, which consists of three \texttt{FC} layers with 256, 128, and 1 neuron(s). To train the vanilla latency predictor, we first sample 1,000 random transformable architecture candidates $\{\overline{\alpha}_i\}_{i=1}^{1,000}$ from the hybrid transformable search space $\mathcal{A}$. After that, we transform each linear \texttt{MBConv} operator with multiple consecutive linear layers into one single \texttt{Conv} layer to derive their corresponding shallow networks. Finally, we deploy the resulting shallow networks on Xavier and Nano to collect their latency measurements. Note that the total design cost, including (1) collecting 1,000 architecture-latency pairs and (2) training the vanilla latency predictor, only takes less than one hour on both Xavier and Nano.

\begin{figure}[t]
    \begin{center}
        \includegraphics[width=1.0\columnwidth]{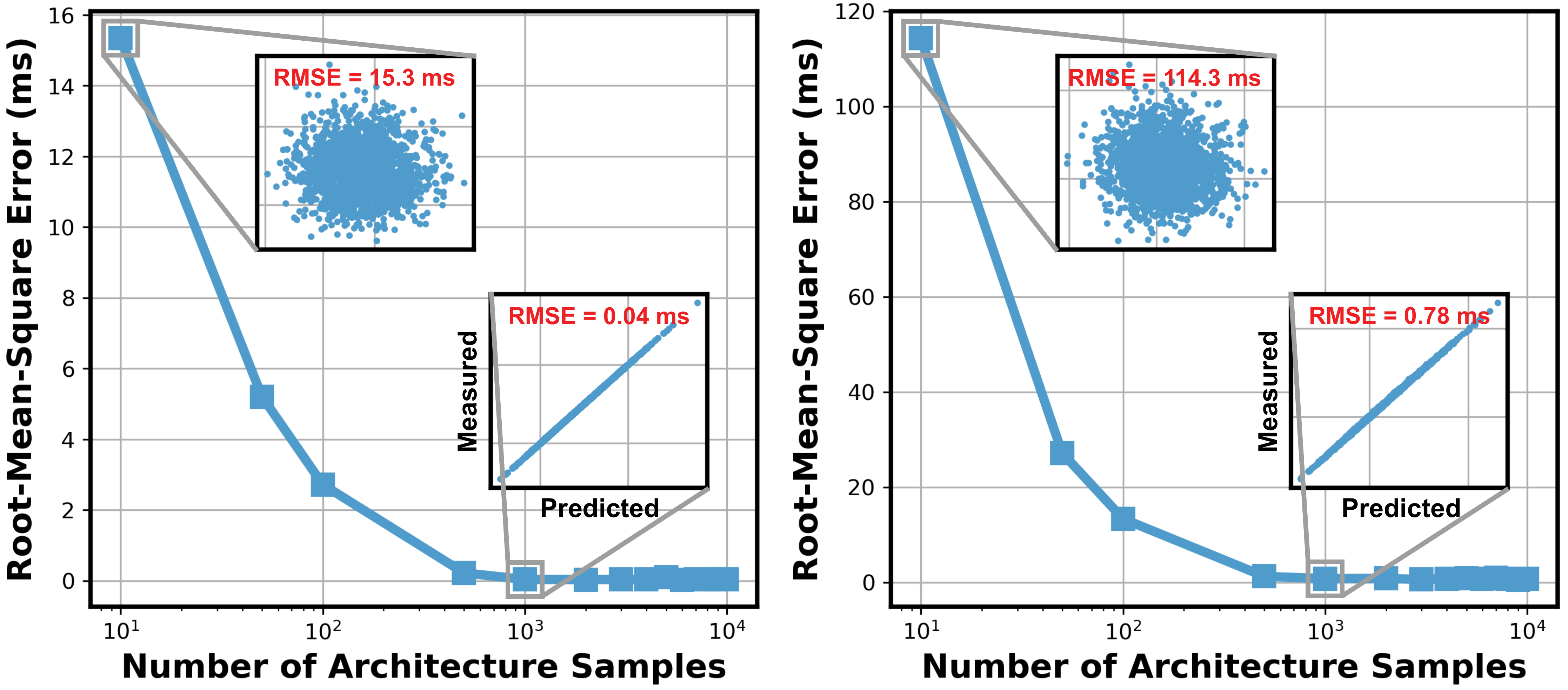}
    \end{center}
    \vspace{-4pt}
    \caption{Latency prediction results on Xavier (\textit{left}) and Nano (\textit{right}).}
    \vspace{-6pt}
    \label{fig:latency-prediction-results}
\end{figure}

\textbf{Results.}
The latency prediction results on the validation set with 1,000 random architecture-latency pairs are illustrated in Fig.~\ref{fig:latency-prediction-results}, which clearly demonstrates that the proposed vanilla latency predictor $LAT(\cdot)$ can achieve an extremely low root-mean-square error (RMSE) of 0.04\,ms on Xavier and 0.78\,ms on Nano, respectively. In addition, we also observe that an increasing number of architecture-latency pairs may further improve the latency prediction performance, in which 1,000 architecture-latency pairs are sufficient to yield reliable latency prediction performance on both Xavier and Nano.

\subsection{Sandwich-Inspired Transformable Search}
\label{sec:sandwich-inspired-transformable-search}

As shown in Eq~(\ref{eq:darts-softmax-relaxation}), DARTS \cite{liu2018darts} simultaneously optimizes all the operator candidates in the supernet \cite{cai2018proxylessnas}, which, despite its efficacy, involves considerable memory consumption during the search process and thus can only be applied to small proxy task (e.g., CIFAR-10). However, the resulting optimal network on small proxy task often exhibits sub-optimal accuracy on target task \cite{cai2018proxylessnas, cai2019once}. To enable proxyless search, GDAS \cite{dong2019searching} further leverages Gumbel-Softmax reparameterization \cite{jang2016gumbel-softmax} to discretize single-path sub-networks from the supernet. Thanks to its memory efficiency, GDAS has been widely followed in subsequent HW-DNAS methods \cite{li2020edd, luo2022you, luo2022lightnas, hu2020tfnas, li2022physics, zhang2021dian}. Specifically, GDAS first relaxes the architecture parameters $\alpha \in \mathbb{R}^{L \times N}$ and then re-formulates Eq~(\ref{eq:darts-softmax-relaxation}) as follows:
\begin{equation}
    x_{l+1} = \sum_{n=1}^{N} \overline{u}_l^n \cdot o_n(x_l), \,\, \mathrm{s.t.}, \,\, u_l^n = \frac{\exp((p_l^n + g_l^n)/\tau)}{\sum\limits_{n'=1}^N \exp((p_l^{n'} + g_l^{n'})/\tau)}
    \label{eq:gdas-gumbel-softmax-relaxation-1}
\end{equation}
where $g \sim \mathrm{Gumbel(0, 1)}$ is the Gumbel distribution \cite{jang2016gumbel-softmax} and $\tau > 0$ is the softmax temperature. Similar to Eq~(\ref{eq:alpha-sparse-matrix}), $\overline{u}$ is the one-hot encoding of $u$, which can be formulated as follows:
\begin{equation}
    \overline{u}_l^n = \mathrm{one\_hot}(u) = \begin{cases}
        1, & \mathrm{if} \,\,\, u_{l}^{n} = \mathop{\mathrm{arg\,max}}_n ||u_{l}|| \\
        0, & \mathrm{otherwise}
    \end{cases}
    \label{eq:u-sparse-matrix}
\end{equation}
As such, the involved memory consumption is significantly reduced from \textit{multi-path} to \textit{single-path} since $x_{l+1}$ only depends on the operator candidate $o_n$ with $\overline{u}_l^n=1$.

\begin{figure}[t]
    \begin{center}
        \includegraphics[width=1.0\columnwidth]{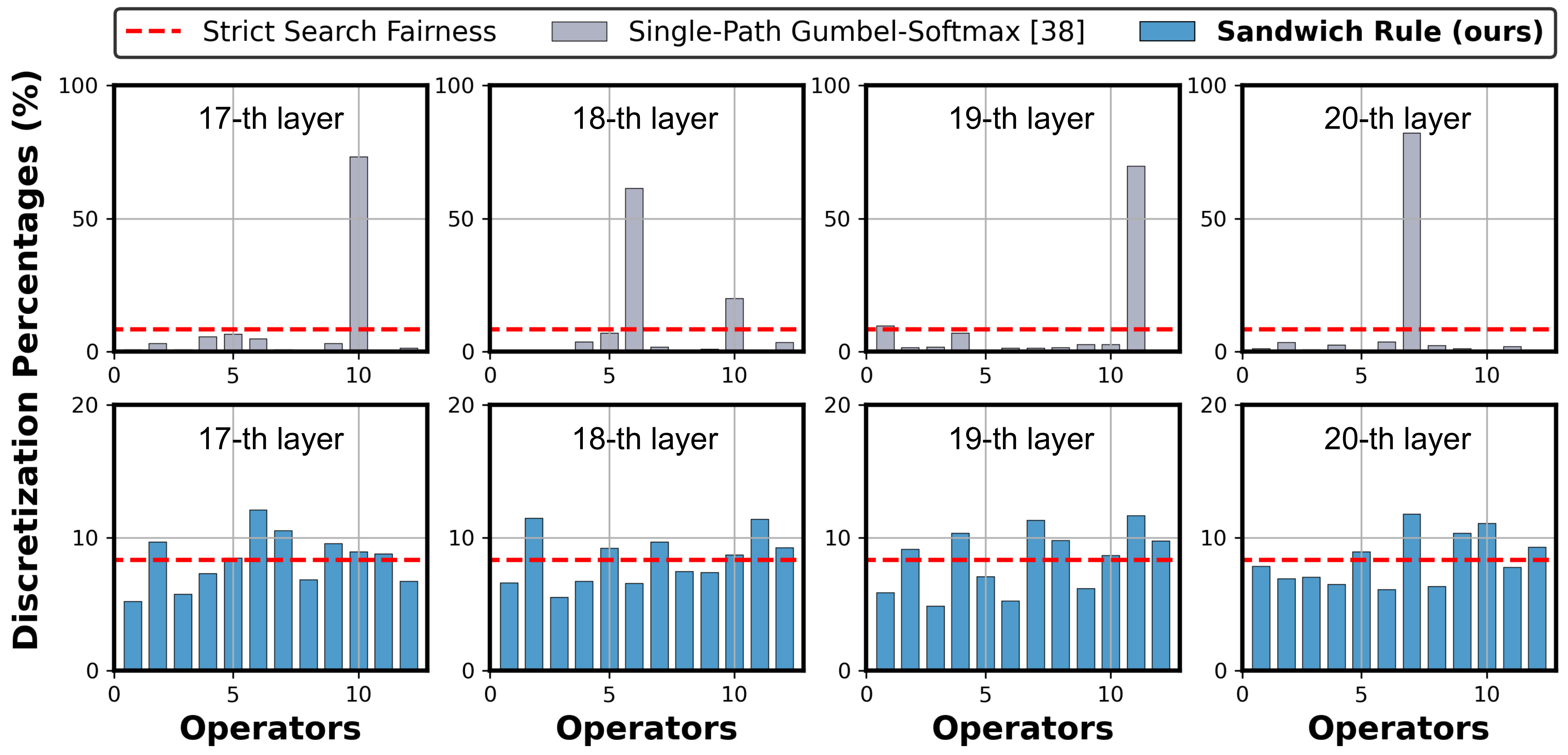}
    \end{center}
    \vspace{-4pt}
    \caption{Operator discretization percentages of single-path Gumbel-Softmax reparameterization (\textit{top}) \cite{dong2019searching} and Gumbel-Softmax Top-$k$ reparameterization \cite{kool2019gumbel-softmax-topk} with the proposed sandwich rule (\textit{bottom}).}
    \vspace{-6pt}
    \label{fig:gumbel-softmax-vs-sandwich}
\end{figure}

\faFrownO\,\,\textbf{The Rich-Get-Richer Collapse.}
The above single-path discretization \cite{dong2019searching}, despite being able to alleviate the memory bottleneck, suffers from significant search bias and thus often ends up with sub-optimal network solutions \cite{chu2021fairnas}. The rationale behind this is that the operators discretized in the early search process are also more likely to be discretized in the subsequent search process (i.e., \textit{the rich get richer}). As a result, the search engine focuses on discretizing a limited number of operators throughout the search process, leaving the remaining operators less explored (see Fig.~\ref{fig:gumbel-softmax-vs-sandwich} (\textit{top})). This is because the proposed hybrid transformable search space features $|\mathcal{O}|=12$ operator candidates, which is $8.2\times10^{4}\approx12^{21}/7^{21}$ larger than recent representative search spaces with $|\mathcal{O}|=7$ operator candidates \cite{cai2018proxylessnas}. In light of this, the above single-path discretization \cite{dong2019searching} may deal with recent representative small search spaces, but cannot generalize to the proposed larger hybrid transformable search space due to the rich-get-richer search collapse.

\faSmileO\,\,\textbf{The Sandwich Rule.}\,To remedy the above rich-get-richer search collapse, we introduce a simple yet effective sandwich rule, which can enforce strong search fairness while incurring comparable search cost. Specifically, we turn to Gumbel-Softmax Top-$k$ reparameterization \cite{kool2019gumbel-softmax-topk}, which can generalize vanilla Gumbel-Softmax reparameterization \cite{jang2016gumbel-softmax} to iteratively discretize $k$ single-path sub-networks without replacement, spanning from the most to the least important single-path sub-network. In practice, after sampling one single-path sub-network, we re-normalize the probability distribution $p$ to exclude the operators sampled by previous single-path sub-networks. After that, we use the re-normalized probability distribution to sample subsequent single-path sub-networks without replacement. In contrast to GDAS \cite{dong2019searching} and its follow-up NAS works \cite{li2020edd, hu2020tfnas, li2022physics, zhang2021dian, luo2022lightnas, luo2022you} that only discretize the most important single-path sub-network (see Eq~(\ref{eq:gdas-gumbel-softmax-relaxation-1})), the proposed sandwich rule focuses on discretizing the following three single-path sub-networks during each search iteration:
\begin{equation}
    w^*(\alpha) = \mathop{\mathrm{arg\,min}}_{w}  \mathcal{L}_{train}^{most}(w, \alpha) + \mathcal{L}_{train}^{rand}(w, \alpha) + \mathcal{L}_{train}^{least}(w, \alpha)
    \label{eq:sandwich-inspired-search}
\end{equation}
where $\mathcal{L}_{train}^{most}(\cdot)$, $\mathcal{L}_{train}^{least}(\cdot)$, and $\mathcal{L}_{train}^{rand}(\cdot)$ denote the training loss functions of the most important single-path sub-network, the least important single-path sub-network, and the random single-path sub-network that excludes the above two single-path sub-networks, respectively.

\textbf{Remarks.}
The above sandwich rule involves three forward propagations and one backward propagation, which marginally increases the search cost from 0.4 GPU-days to 0.6 GPU-days compared with the single-path discretization (see Eq~(\ref{eq:gdas-gumbel-softmax-relaxation-1})). In parallel, compared with another natural option that discretizes $|\mathcal{O}|=12$ single-path sub-networks to enforce strict search fairness, the above sandwich rule can maintain considerable search efficiency (i.e., 0.6 GPU-days vs. 2.1 GPU-days), while achieving similar search fairness (see Fig.~\ref{fig:gumbel-softmax-vs-sandwich} (\textit{bottom})). An ablation study is also provided in TABLE~\ref{tab:ablation-of-sandwich}.

\begin{figure*}[t]
    \begin{center}
        \includegraphics[width=1.0\linewidth]{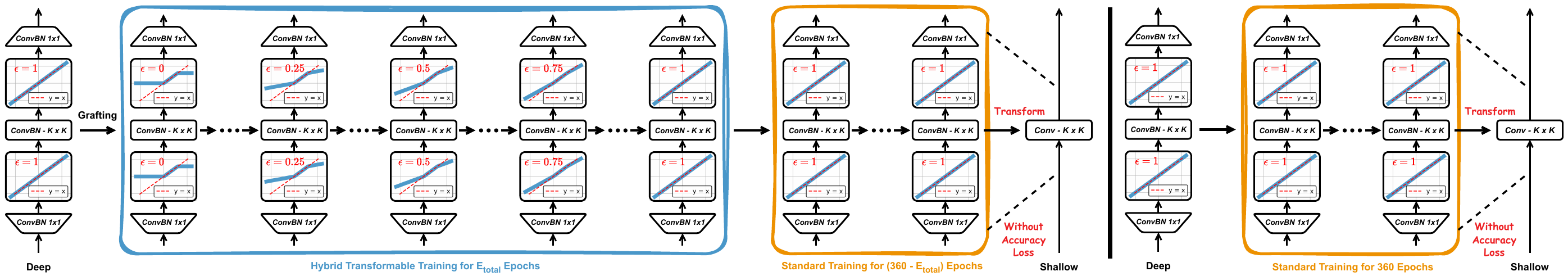}
    \end{center}
    \vspace{-4pt}
    \caption{An intuitive overview of the proposed hybrid transformable training technique (\textit{left}) and the standard training counterpart (\textit{right}).}
    \vspace{-6pt}
    \label{fig:overview-htt}
\end{figure*}

\subsection{Hardware-Aware Transformable Search}
\label{sec:hardware-aware-transformable-search}

To enable hardware-aware transformable search, we further incorporate the proposed vanilla latency predictor $LAT(\cdot)$ into the proposed sandwich-inspired differentiable search algorithm to explore the optimal transformable network around the specified latency constraint, which can be formulated as follows:
\begin{equation}
    \mathop{\mathrm{minimize}}\limits_{\alpha} \, \mathcal{L}_{valid}(w^*(\alpha), \alpha) + \lambda \cdot \left[\frac{LAT(\alpha)}{T} - 1\right]
    \label{eq:hardware-aware-search}
\end{equation}
where $\lambda$ is the coefficient to control the trade-off magnitude and $T$ is the specified latency constraint. In contrast to previous well-established HW-DNAS conventions \cite{cai2018proxylessnas,wu2019fbnet, stamoulis2019single}, $\lambda$ is not a constant but a learnable hyper-parameter, which can be automatically optimized during the search process to guarantee $LAT(\alpha)=T$. For simplicity of notation, we use $\mathcal{L}(w, \alpha, \lambda)$ to denote the search optimization objected in Eq~(\ref{eq:hardware-aware-search}). Finally, we introduce gradient descent to optimize both $w$ and $\alpha$ \cite{liu2018darts}, whereas $\lambda$ is optimized using gradient ascent as follows:
\begin{equation}
    \begin{cases}
        w^* = w - \eta_w \cdot \frac{\partial \mathcal{L}(w, \alpha, \lambda)}{\partial w}, \,
        \alpha^* = \alpha - \eta_{\alpha} \cdot \frac{\partial \mathcal{L}(w, \alpha, \lambda)}{\partial \alpha} \\
        \lambda^* = \lambda + \eta_{\lambda} \cdot \frac{\partial \mathcal{L}(w, \alpha, \lambda)}{\partial \lambda} = \lambda + \eta_{\lambda} \cdot \left[\frac{LAT(\alpha)}{T} - 1\right]
    \end{cases}
    \label{eq:hardware-aware-update}
\end{equation}
where $\eta_w$, $\eta_{\alpha}$, and $\eta_{\lambda}$ are the learning rates of $w$, $\alpha$, and $\lambda$, respectively. Below we further discuss the rationale behind Eq~(\ref{eq:hardware-aware-search})-(\ref{eq:hardware-aware-update}). As illustrated in Fig.~\ref{fig:lambda-tradeoff}, a smaller $\lambda$ ends up with the architecture with high latency and vice versa for a larger $\lambda$. In light of this observation, if $LAT(\alpha)<T$, the gradient ascent scheme decreases $\lambda$ to diminish the trade-off magnitude. As a result, $LAT(\alpha)$ tends to increase towards $T$ in the next search iteration. Likewise, if $LAT(\alpha)>T$, the gradient ascent scheme increases $\lambda$ to enhance the trade-off magnitude. As a result, $LAT(\alpha)$ tends to decrease towards $T$ in the next search iteration. Finally, the search process converges towards $LAT(\alpha)=T$, which not only satisfies the specified latency constraint $T$ but also maintains optimized accuracy.

{\large{\faLightbulbO}}\,
\textbf{Differentiable Analysis.}
As shown in \cite{liu2018darts, dong2019searching, wu2019fbnet} and Eq~(\ref{eq:hardware-aware-update}), $\mathcal{L}(w, \alpha, \lambda)$ is differentiable with respect to both $w$ and $\lambda$, where only continuous transformations are involved. Furthermore, we discuss the differentiability of $\alpha$ as follows:
\begin{equation}
    \begin{aligned}
        &\frac{\partial \mathcal{L}(w, \alpha, \lambda)}{\partial \alpha} = \frac{\partial \mathcal{L}_{valid}(w^*(\alpha), \alpha)}{\partial \alpha} + \frac{\lambda}{T} \cdot \frac{\partial LAT(\alpha)}{\alpha} \\
        & = \frac{\mathcal{L}_{valid}}{\partial \overline{u}} \cdot \frac{\partial \overline{u}}{\partial u} \cdot \frac{\partial u}{\partial p} \cdot \frac{\partial p}{\partial \alpha} + \frac{\lambda}{T} \cdot \frac{\partial LAT(\alpha)}{\partial \overline{u}} \cdot \frac{\partial \overline{u}}{\partial u} \cdot \frac{\partial u}{\partial p} \cdot \frac{\partial p}{\partial \alpha}
    \end{aligned}
    \label{eq:differentiable-analysis-alpha}
\end{equation}
where $\frac{\partial \overline{u}}{\partial u} \approx 1$ as discussed in \cite{bengio2013estimating}. In addition, $\frac{\partial LAT(\alpha)}{\partial \overline{u}}$ is determined by the vanilla latency predictor $LAT(\cdot)$, which can be interpreted through a one-time backward propagation. Apart from these, other terms are apparently differentiable since only continuous transformations are involved \cite{liu2018darts, dong2019searching, wu2019fbnet}.

\subsection{Hybrid Transformable Training}
\label{sec:hybrid-transformable-training}

We next introduce the proposed hybrid transformable training technique to train the searched transformable network to unleash its attainable accuracy. Specifically, instead of directly training the searched transformable network with both linear and non-linear \texttt{MBConv} operators, we can also (1) start from the same network where all the \texttt{MBConv} operators are non-linear and (2) then gradually eliminate its non-linearity during the early training process to convert it back to the original transformable network. An intuitive comparison between the proposed hybrid transformable training technique and the standard training counterpart is visualized in Fig.~\ref{fig:overview-htt}.

Without loss of generality, we use $Act(\cdot)$ to represent the default non-linear activation. For the searched transformable network with both linear and non-linear \texttt{MBConv} operators, we first graft the linear activation in all its linear \texttt{MBConv} operators with the following hybrid activation $Act^*(\cdot)$:
\begin{equation}
    Act^*(x) = Act(x) + \epsilon \cdot \left[x - Act(x)\right]
    \label{eq:linearity-removal}
\end{equation}
where $x$ is the input data and $\epsilon \in [0, 1]$ is the hyper-parameter to balance the linearity and non-linearity. As shown in Fig.~\ref{fig:overview-htt}, $Act^*(\cdot)$ is strictly linear when $\epsilon=1$ and non-linear when $\epsilon\neq1$. To this end, we set $\epsilon$ as follows to enable progressive non-linearity elimination during the early training process:
\begin{equation}
    \epsilon = 
    \begin{cases}
    E_{curr}/E_{total}, & \mathrm{if}\,\,\, E_{curr} < E_{total} \\
    1, & \mathrm{otherwise}
    \end{cases}
    \label{eq:epsilon}
\end{equation}
where $E_{curr}$ is the current training epoch and $E_{total}$ is the total number of training epochs for progressive non-linearity elimination. An example is provided in Fig.~\ref{fig:overview-htt}. As visualized in Fig.~\ref{fig:overview-htt}, at the beginning of training (i.e., $E_{curr}=0$), we have $Act^*(x)=Act(x)$, which is equivalent to the default non-linear activation $Act(\cdot)$ and thus maintains strong non-linearity. As the training proceeds, the grafted non-linearity is gradually eliminated. Finally, when $E_{curr} \geq E_{total}$, we have $Act^(x)=x$, which is equivalent to the standard linear identity mapping activation and thus converts the grafted non-linear \texttt{MBConv} operator back to its linear counterpart. This reveals that the searched transformable network, when properly engineered, can be seamlessly enhanced with grafted non-linearity during the early training epochs, which can largely benefit the remaining training epochs towards better training accuracy \cite{chen2023vanillanet}. More importantly, the network structure remains the same at the end of training, allowing us to unleash the attainable accuracy of the searched transformable network without degrading its on-device efficiency.

\subsection{Hybrid Arbitrary-Resolution Training}
\label{sec:hybrid-any-resolution-elastic-training}

\begin{figure}[t]
    \begin{center}
        \includegraphics[width=1.0\columnwidth]{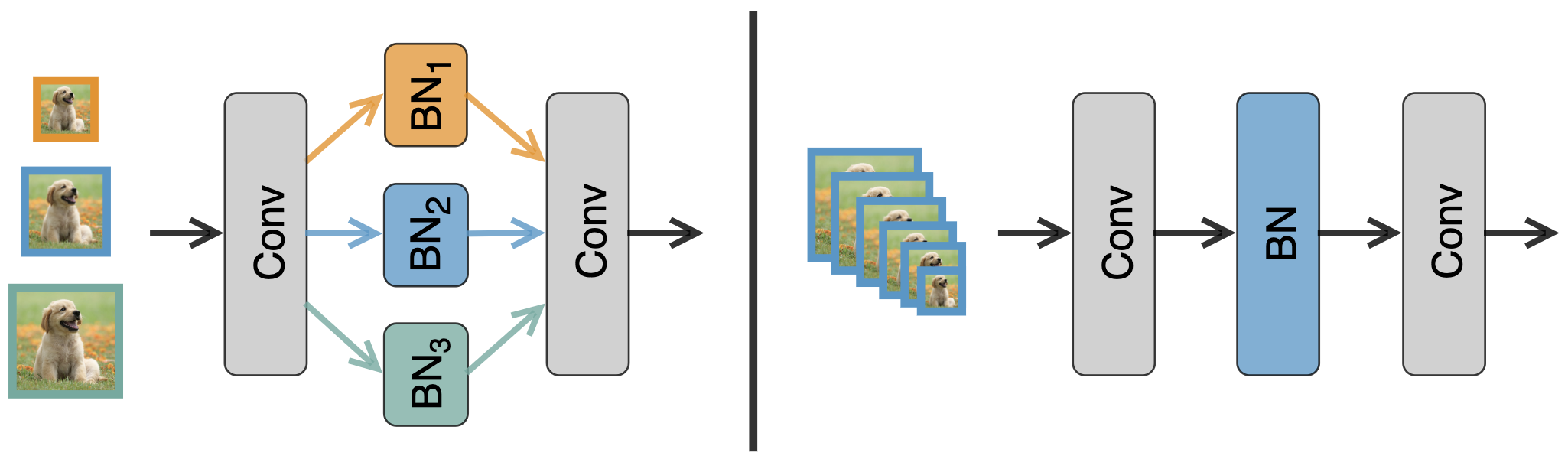}
    \end{center}
    \vspace{-4pt}
    \caption{RS-Net \cite{wang2020resolution} (\textit{left}) vs. our arbitrary-resolution training (\textit{right}).}
    \vspace{-6pt}
    \label{fig:comparisons-overview-rsnet-ours}
\end{figure}

Furthermore, previous SOTA HW-DNAS methods \cite{li2020edd, zhang2021dian, luo2022lightnas, cai2018proxylessnas, wu2019fbnet, stamoulis2019single, hu2020tfnas, fang2020densely, luo2022you, li2022physics} focus on training the searched networks using the standard training recipe. However, the well-trained networks can only accommodate static computational budgets \cite{yu2018slimmable, yu2019universally, wang2018skipnet, yang2022once, wang2020resolution} and cannot naturally enable elastic/dynamic accuracy-efficiency trade-offs during runtime executions on target hardware \cite{luo2024efficient}. Nonetheless, the natural neural network elasticity during inference is of paramount importance for real-world embedded environments, in which the available computational resources may change over time \cite{luo2024efficient}. For example, in battery-powered embedded environments (e.g., mobile phones), the battery level dynamically changes. As a result, the available computational resources decrease as the battery level decreases. To address this issue, RS-Net \cite{wang2020resolution} introduces switchable batch normalization \cite{yu2018slimmable} to eliminate the train-test resolution discrepancy, which tailors an independent batch normalization for each input resolution during runtime resolution switching (see Fig.~\ref{fig:comparisons-overview-rsnet-ours}). In contrast to \cite{yu2018slimmable, yu2019universally, wang2018skipnet, yang2022once}, RS-Net does not introduce additional overheads during runtime resolution switching. However, RS-Net can only support limited input resolutions since each input resolution must come with one tailored batch normalization \cite{yu2018slimmable, yu2019universally}.

\textbf{Arbitrary-Resolution Elastic Training.}
To overcome such limitations, we introduce an effective arbitrary-resolution elastic training to enable runtime resolution switching across arbitrary input resolutions. Without loss of generality, we consider the cross-entry training loss function $\mathcal{L}_{ce}(W, x_i^{224}, y_i)$, where $W$ is the network weight and $(x_i^{224}, y_i)$ is the input-target pair with the default input resolution of $224\times224$. Finally, we can formulate the standard training process as follows:
\begin{equation}
    W_{t+1} = W_{t} - \eta_t \cdot \frac{1}{B} \sum_{i \in \mathcal{B}(t)} \frac{\partial \mathcal{L}_{ce}(W_t, x_i^{224}, y_i)}{\partial W_t}
    \label{eq:sgd-update}    
\end{equation}
where $\eta_t$ is the learning rate and $\mathcal{B}(t)$ is the random sampler to generate an input batch of $B$ input-target pairs $\{(x_i^{224}, y_i)\}_{i=1}^B$ during the training iteration $t$. Nonetheless, the above standard training process can only optimize the given network under the fixed input resolution of $224\times224$ and cannot generalize to unseen input resolutions (see Observation~V) due to the train-test resolution discrepancy \cite{touvron2019fixing, wang2020resolution}. To tackle this issue, we further introduce multiple input batches with different input resolutions and re-formulate Eq~(\ref{eq:sgd-elastic-update}) as follows:
\begin{equation}
    W_{t+1} = W_{t} - \eta_t \cdot \frac{1}{M} \cdot \frac{1}{B} \sum_{m=1}^{M} \sum_{i \in \mathcal{B}(t)} \frac{\partial \mathcal{L}_{ce}(W_t, \mathcal{P}_m(x_i^{224}), y_i)}{\partial W_t}
    \label{eq:sgd-elastic-update}    
\end{equation}
where $\mathcal{P}_m(\cdot)$ is the pre-defined policy to transform the default input resolution of $224\times224$ to target input resolutions and $M$ is the total number of pre-defined policies.

\begin{figure}[t]
    \begin{center}
        \includegraphics[width=1.0\columnwidth]{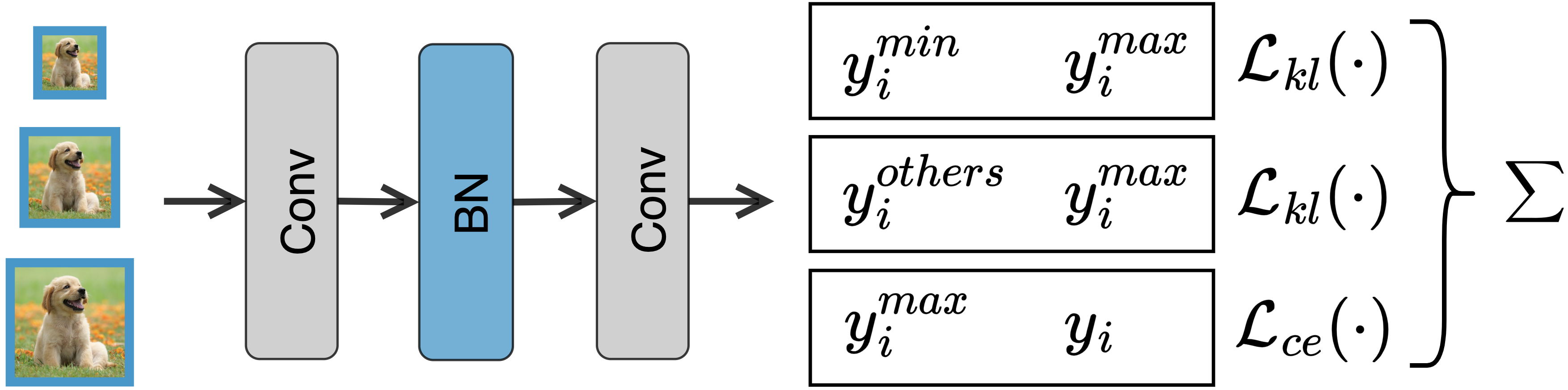}
    \end{center}
    \vspace{-4pt}
    \caption{Illustration of the proposed cross-resolution distillation, in which we consider three input resolutions for the sake of simplicity.}
    \vspace{-6pt}
    \label{fig:overview-cross-resolution-distillation}
\end{figure}

\textbf{Cross-Resolution Distillation.}
As shown in Eq~(\ref{eq:sgd-elastic-update}), the largest input resolution (i.e., $224\times224$) is always optimized during each training iteration $t$. This allows us to leverage the output from the largest input resolution to enhance the training process of subsequent smaller input resolutions towards better training accuracy, and more importantly, without incurring additional training costs. As shown in Fig.~\ref{fig:overview-cross-resolution-distillation}, only the largest input resolution can see the ground-truth target during each training iteration, whereas other smaller input resolutions are optimized to match the output from the largest input resolution. Taken together, we can re-formulate Eq~(\ref{eq:sgd-elastic-update}) as follows:
\begin{equation}
    \begin{aligned}
    & W_{t+1} = W_{t} - \eta_t \cdot \frac{1}{B} \cdot \sum_{i \in \mathcal{B}(t)} \frac{\partial \mathcal{L}_{ce}(W_t, x_i^{224}, y_i)}{\partial W_t} \\ 
    &- \eta_t \cdot \frac{1}{M-1} \cdot \frac{1}{B} \cdot \sum_{m=1}^{M-1} \sum_{i \in \mathcal{B}(t)} \frac{\partial \mathcal{L}_{kl}(W_t, \mathcal{P}_m(x_i^{224}), y_i^{224})}{\partial W_t}
    \end{aligned}
    \label{eq:inplace-distillation}    
\end{equation}
where $\mathcal{L}_{kl}(\cdot)$ is the Kullback–Leibler divergence loss and $y_i^{224}$ is the output from the largest input resolution $x_i^{224}$.

\begin{figure}[t]
    \begin{center}
        \includegraphics[width=1.0\columnwidth]{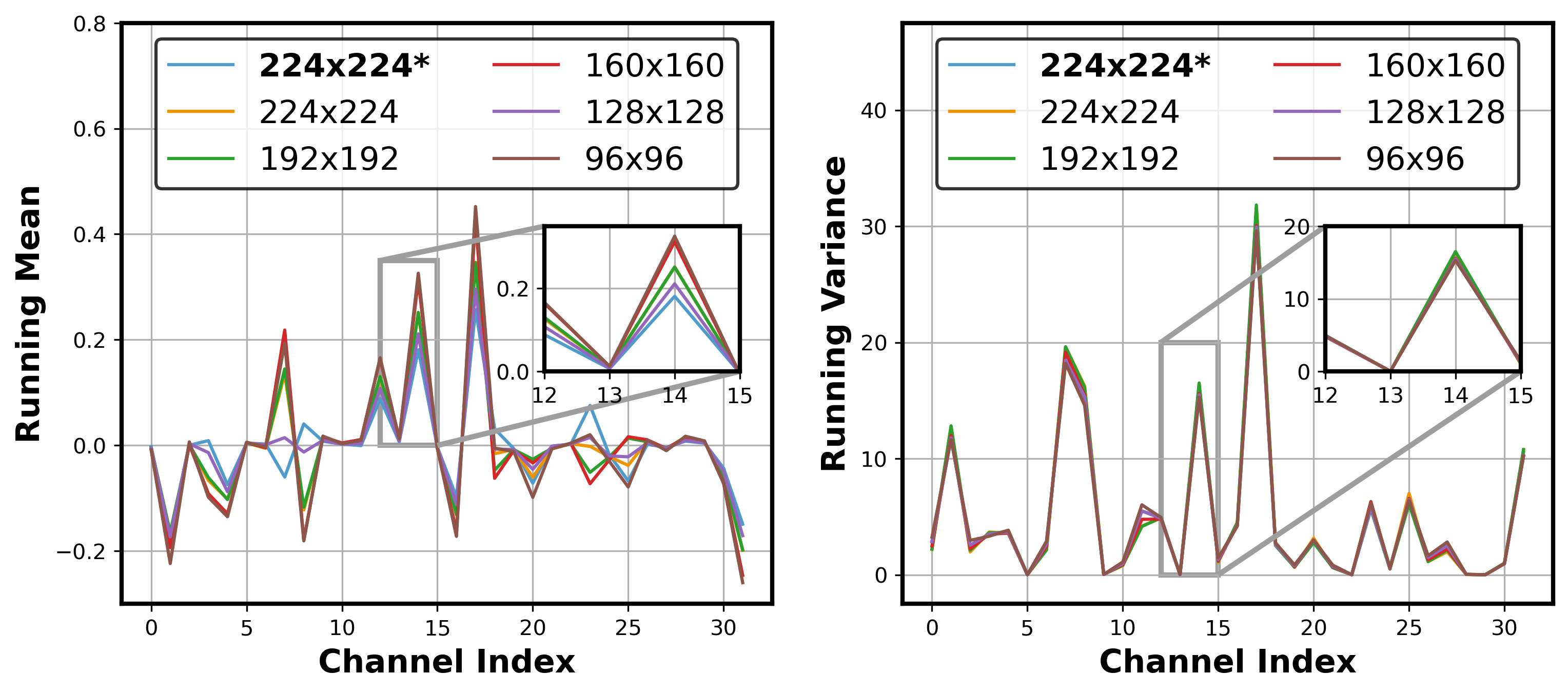}
    \end{center}
    \vspace{-4pt}
    \caption{Visualization of the channel-wise running statistics of the first batch normalization layer of MobileNetV2 \cite{sandler2018mobilenetv2}. $*$ denotes the default baseline without post-training statistics calibration on ImageNet-100.}
    \vspace{-6pt}
    \label{fig:mobilenetv2-bn-calibration-imagenet100}
\end{figure}

\textbf{Post-Training Statistics Calibration.}
As discussed in \cite{hoffer2019mix}, batch normalization layers introduce considerable discrepancy between training and inference due to the inconsistent running mean and variance statistics. For example, as shown in Fig.~\ref{fig:mobilenetv2-bn-calibration-imagenet100}, the running variance statistics are largely consistent across different input resolutions, whereas the running mean statistics may significantly differ. To further eliminate such inconsistency, we turn back to post-training statistics calibration \cite{hoffer2019mix}, which accumulates the running mean and variance statistics for each input resolution at the end of training. In contrast to RS-Net \cite{wang2020resolution} that features switchable batch normalization with multiple batch normalization layers, we still consider one single batch normalization layer during training, which is shared among all the input resolutions (see Fig.~\ref{fig:comparisons-overview-rsnet-ours}). Specifically, for the given well-trained network, we first freeze its trainable parameters and then accumulate the running statistics of its batch normalization layers for each input resolution. More importantly, the above post-training statistics calibration can be conducted offline (i.e., one-time cost) and only requires inference without computation-intensive backward propagation, which takes seconds on one GeForce RTX 3090 GPU.

\textbf{Remarks.}
In this work, we consider $M=17$ different input resolutions, which range from $96\times96$ to $224\times224$ with an interval of $8\times8$. To this end, one natural training option is to simultaneously optimize all the possible input resolutions during each training iteration, which, however, suffers from prohibitive training overheads. To overcome such limitations, we also feature the sandwich rule and simultaneously optimize three different input resolutions during each training iteration, including the largest input resolution, the smallest input resolution, and the random input resolution that excludes the largest and smallest input resolutions, respectively.

\subsection{Linear Transformation}
\label{sec:linear-transformation}

Finally, for the well-trained deep-to-shallow transformable network, we transform each of its linear \texttt{MBConv} operators with multiple consecutive linear layers into one single \texttt{Conv} layer to derive its shallow counterpart, which can boost the on-device efficiency without accuracy loss. Note that transforming one \texttt{Conv} layer and its subsequent \texttt{BN} or linear activation layer is equivalent to linearly manipulating the \texttt{Conv} layer itself. In light of this, below we discuss how to transform multiple consecutive \texttt{Conv} layers into one single \texttt{Conv} layer.

Without loss of generality, we consider two consecutive \texttt{Conv} layers with an input feature $X \in \mathbb{R}^{C_1 \times H_1 \times W_1}$, an intermediate feature $Y \in \mathbb{R}^{C_2 \times H_2 \times W_2}$, and an output feature $Z \in \mathbb{R}^{C_3 \times H_3 \times W_3}$. In addition, $W^{1} \in \mathbb{R}^{C_1 \times C_2 \times K_1 \times K_1}$ and $W^{2} \in \mathbb{R}^{C_2 \times C_3 \times K_2 \times K_2}$ are the learnable weights of the above two consecutive \texttt{Conv} layers. Finally, we can formulate the above two consecutive \texttt{Conv} layers as follows:
\begin{equation}
    \begin{aligned}
        Y_{c,h,w} = \sum_{c_i=0}^{C_1-1} \sum_{k_h^1=0}^{K_1-1} \sum_{k_w^1=0}^{K_1-1} W^{1}_{c_i, c, k_h^1, k_w^1} X_{c_i, h+\Delta k_h^1, w+\Delta k_w^1} \\ 
        Z_{c,h,w} = \sum_{c_i=1}^{C_2-1} \sum_{k_h^2=0}^{K_2-1} \sum_{k_w^2=0}^{K_2} W^{2}_{c_i, c, k_h^2, k_w^2} Y_{c_i, h+\Delta k_h^2, w+\Delta k_w^2} 
    \end{aligned}
    \label{eq:two-conv-layers}
\end{equation}
where $\Delta k_h^1 = k_h^1 - [\frac{K_1 - 1}{2}]$, $\Delta k_w^1 = k_w^1 - [\frac{K_1 - 1}{2}]$, $\Delta k_h^2 = k_h^2 - [\frac{K_2 - 1}{2}]$, and $\Delta k_w^2 = k_w^2 - [\frac{K_2 - 1}{2}]$. Taken together, we can transform the above two consecutive \texttt{Conv} layers as follows:
\begin{equation}
    Z_{c, h, w} = \sum_{c_i = 0}^{C_1 - 1} \sum_{k_h^* = 0}^{K^*- 1} \sum_{k_w^* = 0}^{K^* - 1} W_{c_i, c, k_h^*, k_w^*}^{*} X_{c_i, h+\Delta k_h^*, w+\Delta k_w^*}
    \label{eq:one-conv-layer-1}
\end{equation}
where $\Delta k_h^* = k_h^* - [\frac{K^* - 1}{2}]$ and $\Delta k_w^* = k_w^* - [\frac{K^* - 1}{2}]$. Besides, $W^*$ is the learnable weight of the transformed \texttt{Conv} layer with the kernel size of $K^*=K_1+K_2-1$ \cite{luo2024pearls, luo2024domino}. As demonstrated in \cite{luo2024pearls, luo2024domino}, $W^*$ can be calculated as follows:
\begin{equation}
    W^{*}_{c_i, c_o, h, w} = \sum_{c_j=0}^{C_2-1} \sum_{k_h=0}^{K_2-1} \sum_{k_w=0}^{K_2-1} W^{1}_{c_i, c_j, k_h, k_w} W^{2}_{c_j, c_o, h-\Delta k_h, w-\Delta k_w}
    \label{eq:one-conv-layer-2}
\end{equation}
where $\Delta k_h = k_h - [\frac{K_2-1}{2}]$ and $\Delta k_w = k_w - [\frac{K_2-1}{2}]$. As shown in Eq~(\ref{eq:two-conv-layers})-(\ref{eq:one-conv-layer-2}), the above two consecutive \texttt{Conv} layers with $W^1 \in \mathbb{R}^{C_1 \times C_2 \times K_1 \times K_1}$ and $W^2 \in \mathbb{R}^{C_2 \times C_3 \times K_2 \times K_2}$ can be equivalently transformed into one single \texttt{Conv} layer with $W^* \in \mathbb{R}^{C_1 \times C_3 \times K^* \times K^*}$, both of which strictly maintain the same output $Z$ (i.e., without accuracy loss).

\begin{figure}[t]
    \begin{center}
        \includegraphics[width=1.0\columnwidth]{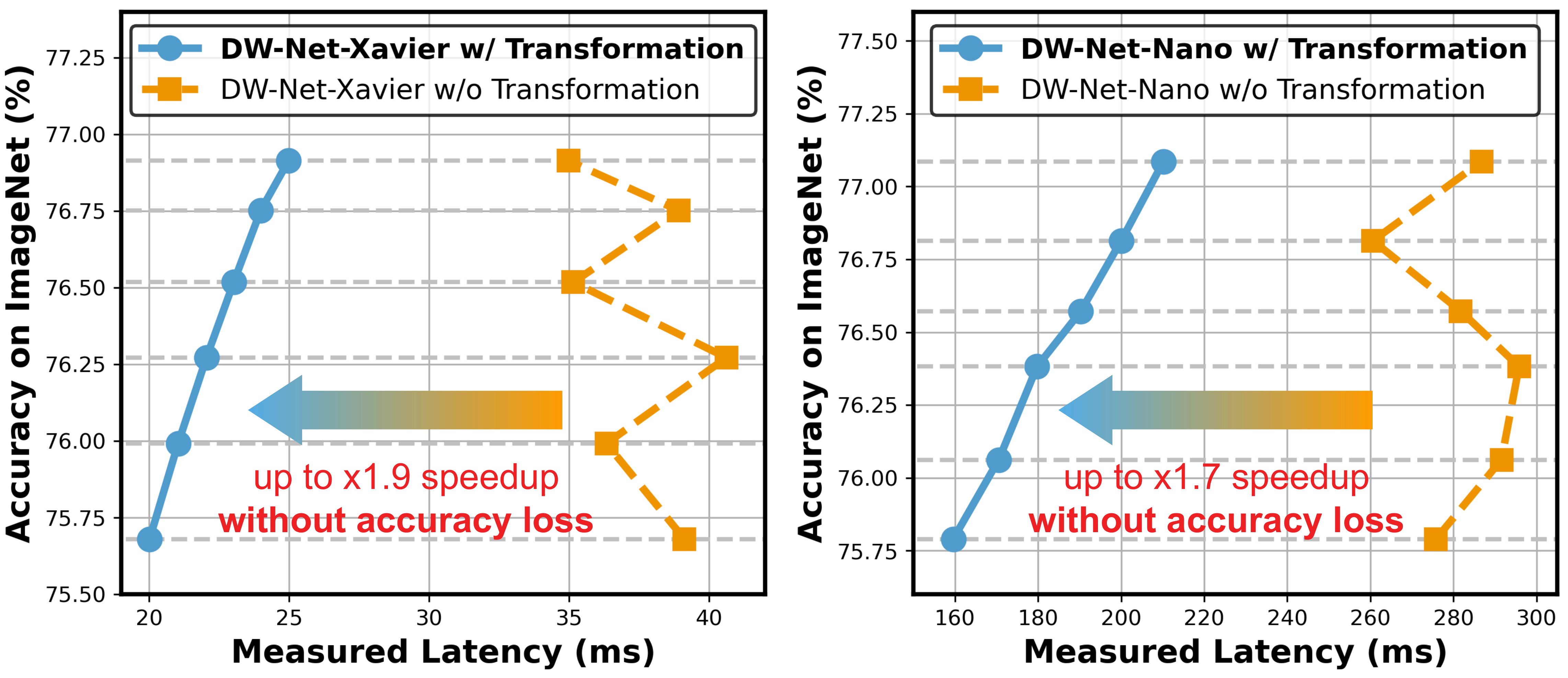}
    \end{center}
    \vspace{-4pt}
    \caption{Linear transformation on Xavier (\textit{left}) and Nano (\textit{right}).}
    \vspace{-6pt}
    \label{fig:effectiveness-transformation}
\end{figure}

\textbf{Remarks and Results.}
To ensure fair comparisons with previous SOTA HW-DNAS methods \cite{wu2019fbnet, cai2018proxylessnas, stamoulis2019single}, we only apply the above linear transformation to linear \texttt{MBConv} operators\footnote{For non-linear \texttt{MBConv} operators, we can also transform the \texttt{Conv} layer and its subsequent \texttt{BN} layer into one \texttt{Conv} layer for better speedups \cite{tensorrt}.}. As shown in Fig.~\ref{fig:effectiveness-transformation}, the above linear transformation achieves up to $\times$1.9 speedups on Xavier and up to $\times$1.7 speedups on Nano without accuracy loss on ImageNet. We also observe that the on-device latency may fluctuate when deep-to-shallow transformation is not applied. In practice, the above fluctuations will be eliminated after deep-to-shallow transformation. The rationale here is that different transformable networks may favor different linear and non-linear \texttt{MBConv} operators in order to maximize the attainable accuracy on target task under the specified computational budgets.

\subsection{Relationships with Previous Methods}
\label{sec:relationships-with-previous-methods}

First and foremost, previous SOTA NAS methods \cite{tan2019mnasnet, howard2019searching, chu2020moga, cai2019once, li2020edd, zhang2021dian, luo2022lightnas, cai2018proxylessnas, wu2019fbnet, stamoulis2019single, hu2020tfnas, fang2020densely, luo2022you, li2022physics} typically feature sufficient network depth in order to explore deep networks with strong accuracy \cite{benmeziane2021comprehensive}. In addition, NetAdaptV2 \cite{yang2021netadaptv2} proposes to merge the network depth and width into one single search dimension to reduce the search cost, whereas D2NAS \cite{lee2024d2nas} and \textit{l}-DARTS \cite{hu2024l} focus on the same cell-based search space and propose different techniques to reduce the search cost (e.g., reduce the network depth during the search process). However, the resulting deep networks do not consist of multiple consecutive linear layers and thus cannot achieve deep-to-shallow transformation. That is, the resulting deep networks, despite their strong accuracy, suffer from poor hardware efficiency since modern hardware processors highly favor shallow networks rather than deep networks \cite{he2015convolutional, goyal2022non}. To overcome such limitations, we may re-engineer previous SOTA NAS methods to explore shallow networks instead. However, the resulting shallow networks, despite their superior hardware efficiency, suffer from inferior accuracy since the attainable accuracy highly relies on sufficient network depth \cite{he2016deep}. This dilemma reveals that previous SOTA NAS methods can only win either accuracy or hardware efficiency and thus cannot enable an aggressive win-win in terms of both.

{\large{\faLightbulbO}}\,
In contrast to previous SOTA NAS methods \cite{tan2019mnasnet, howard2019searching, chu2020moga, cai2019once, li2020edd, zhang2021dian, luo2022lightnas, cai2018proxylessnas, wu2019fbnet, stamoulis2019single, hu2020tfnas, fang2020densely, luo2022you, li2022physics}, DW-NAS opens up a fresh NAS perspective, which pioneers to draw insights from deep-to-shallow transformable networks and establishes the first deep-to-shallow transformable HW-DNAS paradigm to marry the best of both deep and shallow networks. Specifically, DW-NAS focuses on exploring deep networks with sufficient network depth to first win superior accuracy, which then can be equivalently transformed into their shallow counterparts to further win considerable hardware efficiency without accuracy loss. More importantly, DW-NAS provides a holistic solution for both designing and training deep-to-shallow transformable networks. In addition to search, we also draw insights from the searched transformable network and introduce two enhanced training techniques, including hybrid transformable training towards better training accuracy and arbitrary-resolution elastic training towards runtime resolution switching across arbitrary input resolutions. These explicitly distinguish DW-NAS from previous well-established NAS conventions as summarized in TABLE~\ref{tab:nas-comparisons}. More importantly, the proposed deep-to-shallow transformable search paradigm can be seamlessly integrated into more recent NAS works \cite{jiang2025score, nasir2025esm, fayyazi2025marco, lv2025situ} to further enhance their search performance. For example, \cite{lv2025situ} can also explore our proposed hybrid transformable search space rather than traditional search spaces, which can discover more efficient shallow networks with superior hardware efficiency than \cite{lv2025situ} itself.

\begin{table}[t]
\setlength{\tabcolsep}{10pt}
\centering
\caption{Comparisons with previous well-established NAS methods.}
\resizebox{1.0\linewidth}{!}{
    \begin{tabular}{lc|c|c|c|c|c|c|c|}
        & \rot{\small \cite{liu2018darts}} 
        & \rot{\small \cite{dong2019searching}} 
        & \rot{\small \cite{tan2019mnasnet}}
        & \rot{\small \cite{cai2019once}}
        & \rot{\small \cite{wu2019fbnet}}
        & \rot{\small \cite{cai2018proxylessnas}}
        & \rot{\small \,\textbf{ours}\,}\\
         \multicolumn{1}{l|}{Differentiable Search} & \cmark & \cmark & \xmark & \xmark & \cmark & \cmark & \cmark \\
         \multicolumn{1}{l|}{Proxyless Search} & \xmark & \xmark\textcolor{black}{/}\cmark & \cmark & \cmark & \cmark & \cmark & \cmark   \\
         \multicolumn{1}{l|}{Latency Optimization} & \xmark & \xmark & \cmark & \cmark &  \cmark &  \cmark &  \cmark \\
         \multicolumn{1}{l|}{Specified Latency} & \xmark & \xmark & \cmark & \cmark & \xmark & \xmark  & \cmark  \\
         \multicolumn{1}{l|}{Transformable Search} & \xmark & \xmark & \xmark & \xmark & \xmark & \xmark  & \cmark   \\
         \multicolumn{1}{l|}{Transformable Training} & \xmark & \xmark & \xmark & \xmark & \xmark & \xmark  & \cmark   \\
        \multicolumn{1}{l|}{Elastic Training} & \xmark & \xmark & \xmark & \xmark & \xmark & \xmark  & \cmark   \\
         \multicolumn{1}{l|}{Search Cost (GPU-days)} & 1 & 0.2/0.4 & 1,667 & 53 & 9 & 8 & 0.6 \\
    \end{tabular}
}
\label{tab:nas-comparisons}
\end{table}

\section{Experiments}
\label{sec:experiments}

In this section, we conduct extensive experiments, including thorough comparisons with previous SOTA NAS methods and insightful ablation studies, to show the efficacy of DW-NAS. Note that the proposed latency predictor is only used for transformable search with fixed resolution of $224\times224$ and batch size of 8. In this work, all the reported latency results are directly measured on Xavier and Nano for fair comparisons.

\subsection{Experimental Setup}
\label{sec:experimental-settings}

\textbf{Datasets.}
Following FBNet \cite{wu2019fbnet}, we consider two representative large-scale datasets, including ImageNet and ImageNet-100. Specifically, ImageNet consists of 1,281,167 training images and 50,000 validation images, all of which are roughly distributed across 1,000 classes. Furthermore, ImageNet-100 is a subset of ImageNet and consists of 100 random classes from ImageNet. Following recent well-established NAS conventions \cite{wu2019fbnet, cai2018proxylessnas, stamoulis2019single}, we set the default input image size to 224$\times$224 for both network training and latency measurements.

\textbf{Embedded Platforms.}
We consider two popular NVIDIA Jetson intelligent embedded systems, including NVIDIA Jetson AGX Xavier and NVIDIA Jetson Nano, which are abbreviated as Xavier and Nano for the sake of simplicity. Specifically, Xavier is equipped with 512-core Volta GPU, 8-core Carmel ARM CPU, 32\,GB LPDDR4x memory, Ubuntu 20.04, JetPack 4.4, and PyTorch 1.8, whereas Nano is equipped with 128-core Maxwell GPU, quad-core Cortex ARM CPU, 4\,GB LPDDR4 memory, Ubuntu 18.04, JetPack 4.4, and PyTorch 1.8. For Xavier, we use its onboard eMMC as the main storage, whereas we use an external microSD for Nano. For both Xavier and Nano, we use the recommended DC Barrel jack to power the board, where the MAXN power model is also manually turned on to maximize the runtime hardware performance. For fair comparisons, we report the measured latency under the same input batch size of 8.

\begin{figure}[t]
    \begin{center}
        \includegraphics[width=1.0\columnwidth]{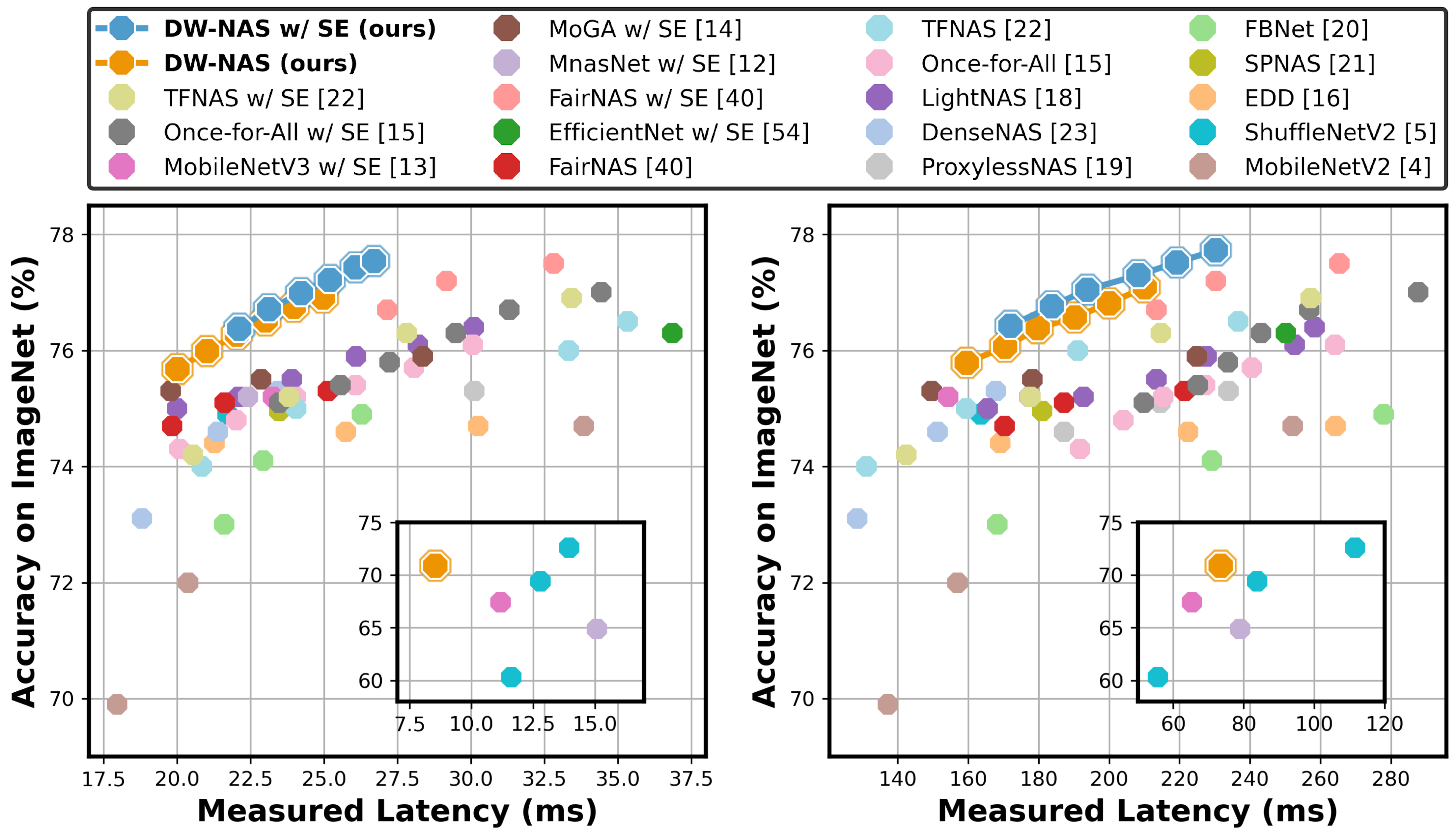}
    \end{center}
    \vspace{-4pt}
    \caption{DW-NAS vs. SOTA efficient networks on Xavier (\textit{left}) and Nano (\textit{right}), where SE is the Squeeze-and-Excitation module \cite{hu2018squeeze}. Note that zoom-in figures compare DW-Net-Small with other SOTA efficient networks under low latency constraints on Xavier and Nano.}
    \vspace{-6pt}
    \label{fig:comparisons-imagenet}
\end{figure}

\textbf{Architecture Search Settings.}
The architecture search experiments of DW-NAS closely follow FBNet \cite{wu2019fbnet}, where both network weights $w$ and architecture parameters $\alpha$ are optimized on ImageNet-100. Specifically, the supernet is trained from scratch for 90 epochs with an input batch size of 192. To stabilize the search process, we also freeze $\alpha$ in the first 10 epochs \cite{wu2019fbnet}. To optimize $w$, we employ SGD with a learning rate of 0.1, a momentum of 0.9, and a weight decay of $3\times10^{-5}$. In parallel, to optimize $\alpha$, we utilize Adam with a learning rate of 0.001 and a weight decay of $1\times10^{-3}$. Furthermore, we initialize the trade-off coefficient $\lambda$ to zero and then optimize $\lambda$ using gradient ascent, where the learning rate is set to 0.0005. Note that all the search experiments are performed on one single GeForce RTX 3090 GPU, each of which takes 0.6 days. Finally, we denote the searched transformable networks on Xavier and Nano as DW-Net-Xavier and DW-Net-Nano, respectively.

\textbf{Architecture Evaluation Settings.}
We also closely follow FBNet \cite{wu2019fbnet} to evaluate the searched transformable networks on ImageNet and ImageNet-100. Specifically, all the searched transformable networks are trained from scratch for 360 epochs with an input batch size of 256 on 4 GeForce RTX 3090 GPUs, where the standard data augmentations are applied throughout this work. The default optimizer is SGD with an initial learning rate of 0.1 (gradually annealed to zero following the cosine schedule), a momentum of 0.9, and a weight decay of $4\times10^{-5}$. Finally, when hybrid transformable training is enabled, the grafted non-linearity is gradually eliminated in the first 120 training epochs (i.e., $E_{total}=120$). Note that the total training epochs still remain 360 regardless of hybrid transformable training as shown in Fig.~\ref{fig:overview-htt} (\textit{left}).

\begin{figure}[t]
    \begin{center}
        \includegraphics[width=1.0\columnwidth]{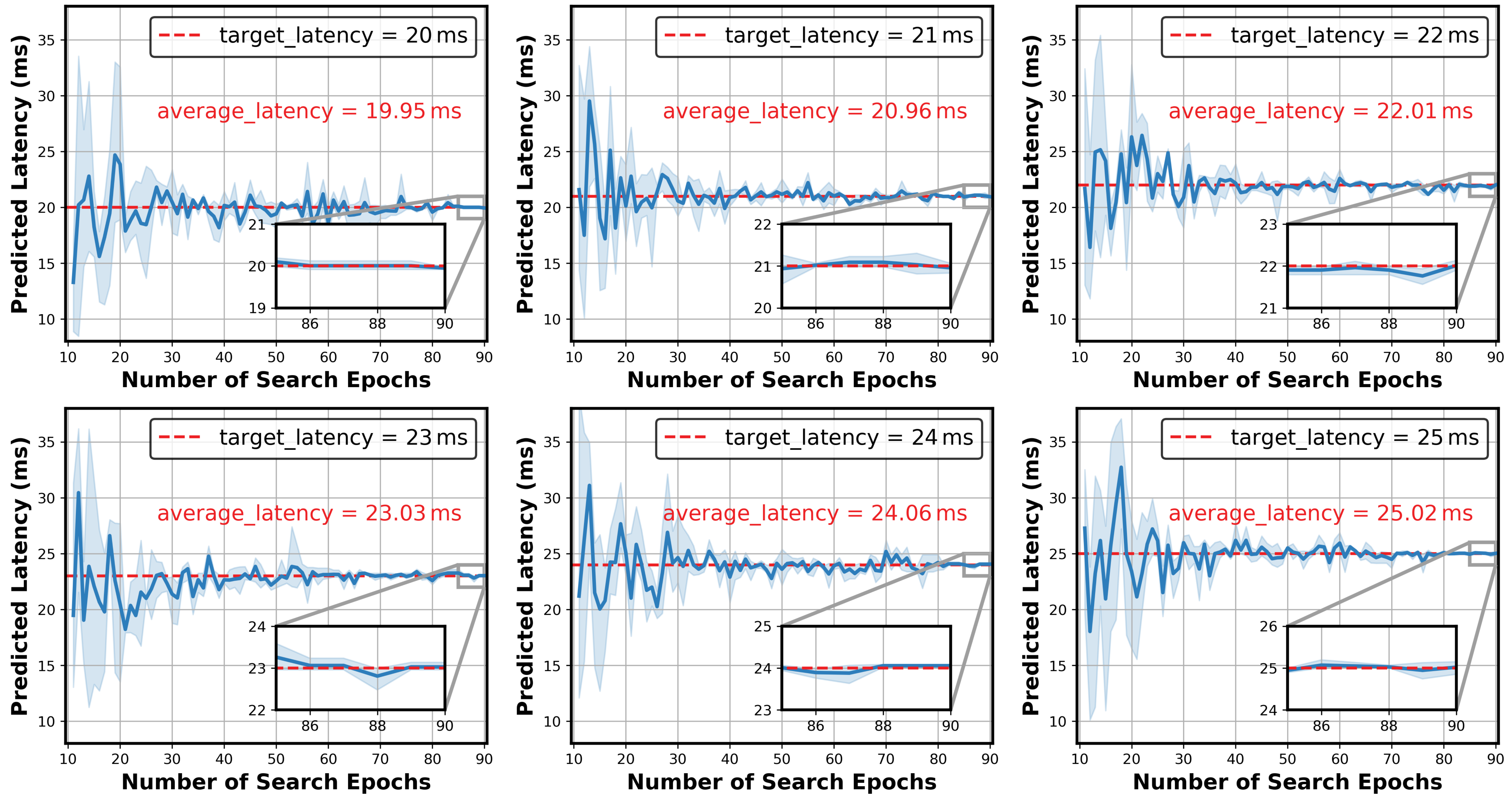}
    \end{center}
    \vspace{-4pt}
    \caption{Visualization of the search process of DW-Net-Xavier under various latency constraints on Xavier ranging from 20\,ms to 25\,ms.}
    \vspace{-6pt}
    \label{fig:xavier-latency-search}
\end{figure}

\subsection{Results and Ablation Studies}
\label{sec:experimental-results}

\textbf{Architecture Evaluation Results.}
To demonstrate the efficacy of DW-NAS, we first compare the searched DW-Nets with a plethora of previous SOTA efficient networks on ImageNet, spanning from \textit{manual} \cite{ma2018shufflenet, sandler2018mobilenetv2} to \textit{automated} \cite{hu2020tfnas, cai2019once, howard2019searching, chu2020moga, tan2019mnasnet, chu2021fairnas, tan2019efficientnet, luo2022lightnas, fang2020densely, cai2018proxylessnas, wu2019fbnet, stamoulis2019single, li2020edd}. For fair comparisons with MobileNetV3-style models, we closely follow previous conventions and introduce the Squeeze-and-Excitation (SE) module \cite{hu2018squeeze}, which serves as an enhancement and can be seamlessly integrated into the searched DW-Nets to achieve better accuracy-efficiency trade-offs. Therefore, we train the searched DW-Nets from scratch under both training settings (i.e., with SE and without SE). The experimental results on ImageNet are illustrated in Fig.~\ref{fig:comparisons-imagenet}, which clearly shows that DW-NAS can achieve significantly better accuracy-efficiency trade-offs than the aforementioned SOTA efficient networks on both Xavier and Nano. For example, compared with FBNet-C \cite{wu2019fbnet}, DW-Net-Xavier-20ms can achieve $+$0.8\% higher accuracy on ImageNet, while at the same time maintaining $\times$1.3 inference speedup on Xavier. Furthermore, we observe that DW-Net-Small\footnote{DW-Net-Small only consists of linear \texttt{MBConv} operators with $K=3$ and $E=6$, which maintains the optimal hardware efficiency upon transformation.} can also achieve better accuracy-efficiency trade-offs than other SOTA efficient networks under extremely low latency constraints on both Xavier and Nano.

\begin{table*}[t]
\centering
\caption{Comparisons with previous relevant NAS methods, in which the reported latency is queried from HW-NAS-Bench \cite{li2021hw} and the remaining metrics are queried from NAS-Bench-201 \cite{dong2020bench}. Similar to EH-DNAS \cite{jiang2021eh}, all the reported results are measured on CIFAR-10.}
\resizebox{1.0\linewidth}{!}{%
\begin{tabular}{cc|cccc|cccc|cccc}
\toprule[0.125em]
\multicolumn{2}{c|}{Approach} &
  \multicolumn{4}{c|}{Google Edge TPU} &
  \multicolumn{4}{c|}{Raspberry Pi 4} &
  \multicolumn{4}{c}{Google Pixel 3} \\ \hline
\multicolumn{1}{c|}{\multirow{2}{*}{\begin{tabular}[c]{@{}c@{}}Search \\ Algorithm\end{tabular}}} &
  \multirow{2}{*}{\begin{tabular}[c]{@{}c@{}}Hardware\\ Feedback\end{tabular}} &
  \multicolumn{1}{c|}{\multirow{2}{*}{\begin{tabular}[c]{@{}c@{}}Latency\\ (ms) $\downarrow$\end{tabular}}} &
  \multicolumn{1}{c|}{\multirow{2}{*}{\begin{tabular}[c]{@{}c@{}}Accuracy\\ (\%) $\uparrow$\end{tabular}}} &
  \multicolumn{1}{c|}{\multirow{2}{*}{\begin{tabular}[c]{@{}c@{}}FLOPs\\ (M) $\downarrow$\end{tabular}}} &
  \multirow{2}{*}{\begin{tabular}[c]{@{}c@{}}Params\\ (M) $\downarrow$\end{tabular}} &
  \multicolumn{1}{c|}{\multirow{2}{*}{\begin{tabular}[c]{@{}c@{}}Latency\\ (ms) $\downarrow$\end{tabular}}} &
  \multicolumn{1}{c|}{\multirow{2}{*}{\begin{tabular}[c]{@{}c@{}}Accuracy\\ (\%) $\uparrow$\end{tabular}}} &
  \multicolumn{1}{c|}{\multirow{2}{*}{\begin{tabular}[c]{@{}c@{}}FLOPs\\ (M) $\downarrow$\end{tabular}}} &
  \multirow{2}{*}{\begin{tabular}[c]{@{}c@{}}Params\\ (M) $\downarrow$\end{tabular}} &
  \multicolumn{1}{c|}{\multirow{2}{*}{\begin{tabular}[c]{@{}c@{}}Latency\\ (ms) $\downarrow$\end{tabular}}} &
  \multicolumn{1}{c|}{\multirow{2}{*}{\begin{tabular}[c]{@{}c@{}}Accuracy\\ (\%) $\uparrow$\end{tabular}}} &
  \multicolumn{1}{c|}{\multirow{2}{*}{\begin{tabular}[c]{@{}c@{}}FLOPs\\ (M) $\downarrow$\end{tabular}}} &
  \multirow{2}{*}{\begin{tabular}[c]{@{}c@{}}Params\\ (M) $\downarrow$\end{tabular}} \\
\multicolumn{1}{c|}{} &
   &
  \multicolumn{1}{c|}{} &
  \multicolumn{1}{c|}{} &
  \multicolumn{1}{c|}{} &
   &
  \multicolumn{1}{c|}{} &
  \multicolumn{1}{c|}{} &
  \multicolumn{1}{c|}{} &
   &
  \multicolumn{1}{c|}{} &
  \multicolumn{1}{c|}{} &
  \multicolumn{1}{c|}{} &
   \\ \hline
\multicolumn{1}{c|}{DARTS (1st) \cite{liu2018darts}} & None &
  \multicolumn{1}{c|}{0.60} &
  \multicolumn{1}{c|}{54.30} &
  \multicolumn{1}{c|}{7.8} &
  0.073 &
  \multicolumn{1}{c|}{3.84} &
  \multicolumn{1}{c|}{54.30} &
  \multicolumn{1}{c|}{7.8} &
  0.073 &
  \multicolumn{1}{c|}{1.61} &
  \multicolumn{1}{c|}{54.30} &
  \multicolumn{1}{c|}{7.8} &
  0.073 \\ \hline
\multicolumn{1}{c|}{DARTS (2nd) \cite{liu2018darts}} & None &
  \multicolumn{1}{c|}{0.60} &
  \multicolumn{1}{c|}{54.30} &
  \multicolumn{1}{c|}{7.8} &
  0.073 &
  \multicolumn{1}{c|}{3.84} &
  \multicolumn{1}{c|}{54.30} &
  \multicolumn{1}{c|}{7.8} &
  0.073 &
  \multicolumn{1}{c|}{1.61} &
  \multicolumn{1}{c|}{54.30} &
  \multicolumn{1}{c|}{7.8} &
  0.073 \\ \hline
\multicolumn{1}{c|}{\multirow{3}{*}{GDAS \cite{dong2019searching}}} &
  None &
  \multicolumn{1}{c|}{1.26} &
  \multicolumn{1}{c|}{93.37} &
  \multicolumn{1}{c|}{153.3} &
  1.073 &
  \multicolumn{1}{c|}{69.19} &
  \multicolumn{1}{c|}{93.37} &
  \multicolumn{1}{c|}{153.3} &
  1.073 &
  \multicolumn{1}{c|}{27.70} &
  \multicolumn{1}{c|}{93.37} &
  \multicolumn{1}{c|}{153.3} &
  1.073 \\ \cline{2-14} 
\multicolumn{1}{c|}{} &
  FBNet LUT \cite{wu2019fbnet} &
  \multicolumn{1}{c|}{1.27} &
  \multicolumn{1}{c|}{93.36} &
  \multicolumn{1}{c|}{153.3} &
  1.073 &
  \multicolumn{1}{c|}{0.01} &
  \multicolumn{1}{c|}{10.00} &
  \multicolumn{1}{c|}{7.8} &
  0.073 &
  \multicolumn{1}{c|}{0.01} &
  \multicolumn{1}{c|}{10.00} &
  \multicolumn{1}{c|}{7.8} &
  0.073 \\ \cline{2-14} 
\multicolumn{1}{c|}{} &
  EH-DNAS \cite{jiang2021eh} &
  \multicolumn{1}{c|}{1.10} &
  \multicolumn{1}{c|}{93.08} &
  \multicolumn{1}{c|}{117.9} &
  0.830 &
  \multicolumn{1}{c|}{56.89} &
  \multicolumn{1}{c|}{93.58} &
  \multicolumn{1}{c|}{121.8} &
  0.858 &
  \multicolumn{1}{c|}{16.97} &
  \multicolumn{1}{c|}{93.58} &
  \multicolumn{1}{c|}{121.8} &
  0.858 \\ \hline
\multicolumn{2}{c|}{\cellcolor{Gray}Double-Win NAS (ours)} &
  \multicolumn{1}{c|}{\cellcolor{Gray}1.04} &
  \multicolumn{1}{c|}{\cellcolor{Gray}93.58} &
  \multicolumn{1}{c|}{\cellcolor{Gray}117.9} &
  \cellcolor{Gray}0.830 &
  \multicolumn{1}{c|}{\cellcolor{Gray}50.98} &
  \multicolumn{1}{c|}{\cellcolor{Gray}93.83} &
  \multicolumn{1}{c|}{\cellcolor{Gray}114.0} &
  \cellcolor{Gray}0.802 &
  \multicolumn{1}{c|}{\cellcolor{Gray}15.43} &
  \multicolumn{1}{c|}{\cellcolor{Gray}93.72} &
  \multicolumn{1}{c|}{\cellcolor{Gray}82.5} &
  \cellcolor{Gray}0.587 \\ 
  \toprule[0.125em]
\end{tabular}%
}
\label{tab:results-on-nas-benchmarks}
\end{table*}

\begin{figure}[t]
    \begin{center}
        \includegraphics[width=1.0\columnwidth]{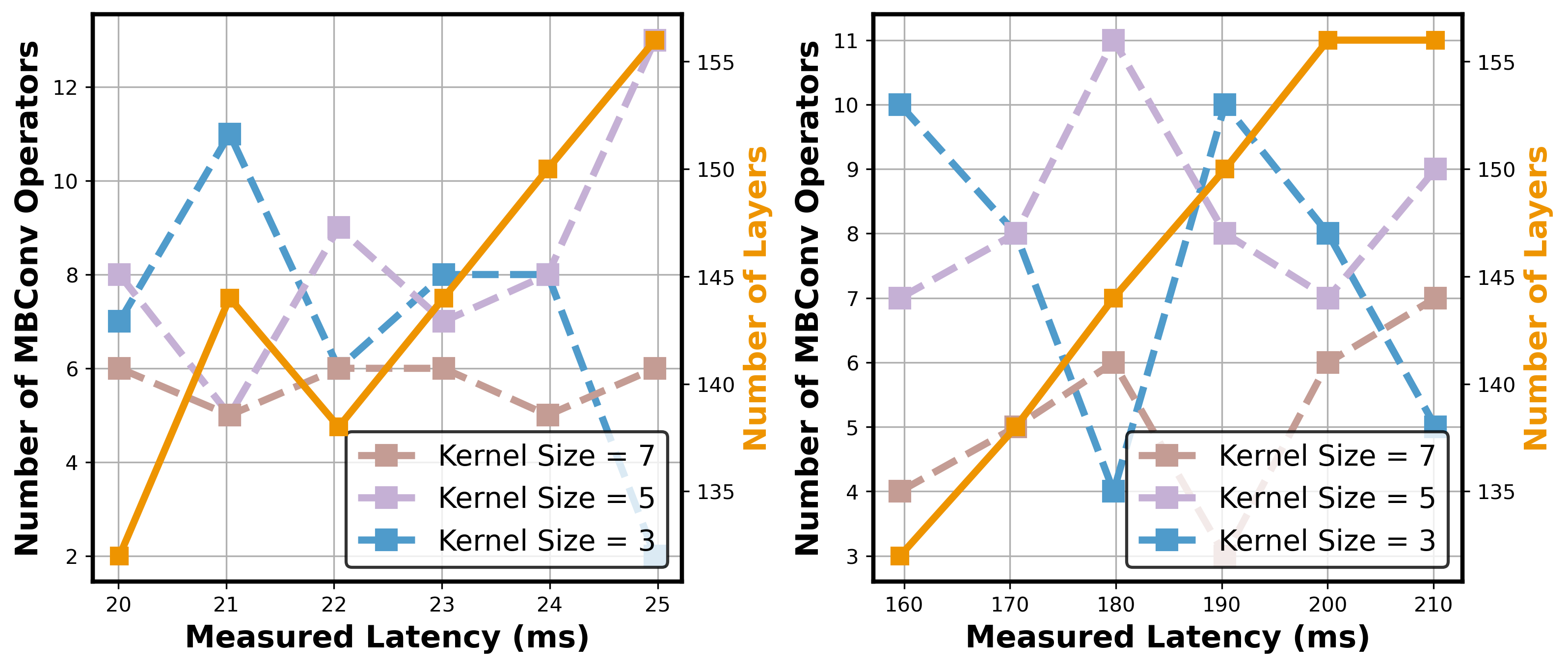}
    \end{center}
    \vspace{-4pt}
    \caption{Architecture search analysis on Xavier\,(\textit{left}) and Nano\,(\textit{right}).}
    \vspace{-6pt}
    \label{fig:architecture-search-analysis}
\end{figure}

\textbf{Architecture Search Analysis.}
As shown in \cite{he2016deep, sandler2018mobilenetv2, cai2018proxylessnas, wu2019fbnet}, there may be various factors that affect the runtime latency, such as the network depth and the kernel size. Therefore, we further analyze the searched transformable networks (i.e., DW-Net-Xavier and DW-Net-Nano) and identify the dominant factor. Specifically, we consider 21 searchable MBConv operators in the search space, where we (1) count the number of layers after deep-to-shallow transformation (i.e., latency vs. depth) and (2) count the number of MBConv operators with kernel sizes of 3, 5, and 7 (i.e., latency vs. kernel size). As illustrated in Fig.~\ref{fig:architecture-search-analysis}, we can easily observe that the runtime latency strongly correlates with the network depth rather than the kernel size. Based on this, we can easily identify the network depth as the dominant factor that affects the runtime latency, which again demonstrates the efficacy of our proposed deep-to-shallow transformable search paradigm.

\textbf{Architecture Search Stability.}
As discussed in Section~\ref{sec:hardware-aware-transformable-search}, DW-NAS strives to navigate the required transformable network around the specified latency constraint in one single search. To investigate its search stability, we further conduct a series of search experiments under various latency constraints, in which we take Xavier as an example. The resulting search process is visualized in Fig.~\ref{fig:xavier-latency-search}, in which each sub-figure is averaged over three search experiments under different random initialization seeds. As shown in Fig.~\ref{fig:xavier-latency-search}, we observe that DW-NAS can maintain strong search stability, which (1) explicitly focuses on exploring possible transformable networks around the specified latency constraint during the search process and (2) consistently navigates the required transformable network around the specified latency constraint at the end of each search experiment.

\textbf{Results on NAS Benchmarks \cite{dong2020bench, li2021hw}.}
Following EH-DNAS \cite{jiang2021eh}, we evaluate DW-NAS on recent well-established NAS benchmarks, including NAS-Bench-201 \cite{dong2020bench} and HW-NAS-Bench \cite{li2021hw}, both of which feature the same cell-based search space that contains 15,625 different cell structures as discussed in \cite{dong2020bench}. To this end, we (1) first replace the hybrid transformable search space with the above cell-based search space and (2) then repeat multiple search experiments to explore top-performing cell structures under various latency constraints. Similar to EH-DNAS, we also consider CIFAR-10 from NAS-Bench-201 and three embedded platforms from HW-NAS-Bench, including Google Edge TPU, Raspberry Pi 4, and Google Pixel 3. The experimental results are summarized in TABLE~\ref{tab:results-on-nas-benchmarks}, which clearly demonstrate that DW-NAS outperforms previous relevant NAS methods in terms of both accuracy and hardware efficiency. For example, compared with EH-DNAS, DW-NAS exhibits $+$0.50\% and $+$0.25\% higher top-1 accuracy on CIFAR-10, which also achieves 1.06$\times$ speedup on Google Edge TPU and 1.12$\times$ speedup on Raspberry Pi 4, respectively. Note that FBNet LUT \cite{wu2019fbnet} may suffer from inferior accuracy on CIFAR-10, especially when targeting Raspberry Pi 4 and Google Pixel 3, due to the inaccurate latency prediction as discussed in \cite{jiang2021eh}.

\begin{figure}[t]
    \begin{center}
        \includegraphics[width=1.0\columnwidth]{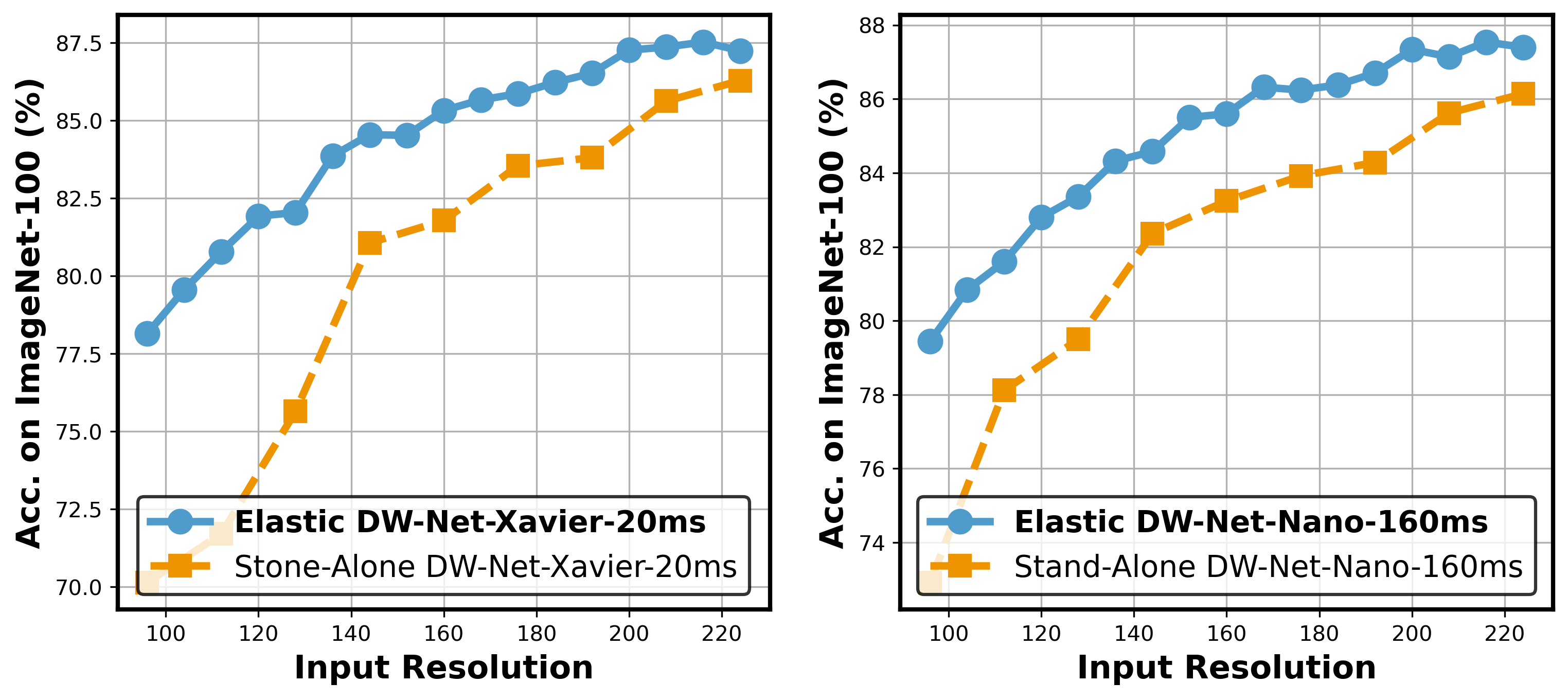}
    \end{center}
    \vspace{-4pt}
    \caption{Arbitrary-resolution elastic training vs. stand-alone training across various input resolutions on Xavier (\textit{left}) and Nano (\textit{right}).}
    \vspace{-6pt}
    \label{fig:elastic-training-against-baselines}
\end{figure}

\textbf{Elastic Training Results.}
As discussed in Section~\ref{sec:hybrid-any-resolution-elastic-training}, we introduce arbitrary-resolution elastic training to enable natural network elasticity towards runtime resolution switching across arbitrary input resolutions. To investigate its efficacy, we further train the searched transformable networks from scratch on ImageNet-100 under different fixed input resolutions, which are denoted as stand-alone networks. As shown in Fig.~\ref{fig:elastic-training-against-baselines}, the proposed arbitrary-resolution elastic training technique can effectively enable natural network elasticity across arbitrary input resolutions. More importantly, the proposed arbitrary-resolution elastic training technique can consistently achieve better accuracy on ImageNet-100 than stand-alone counterparts, especially under small input resolutions (e.g., $96\times96$). More importantly, the cost (e.g., time and memory) of arbitrary-resolution training is similar to standard training. Taking DW-Net-Xavier-20ms as an example, the arbitrary-resolution training requires 15 hours and 25.3 GB memory on 4 GeForce RTX 3090 GPUs, whereas the standard training requires 12 hours and 23.7 GB memory.

\begin{figure}[t]
    \begin{center}
        \includegraphics[width=1.0\columnwidth]{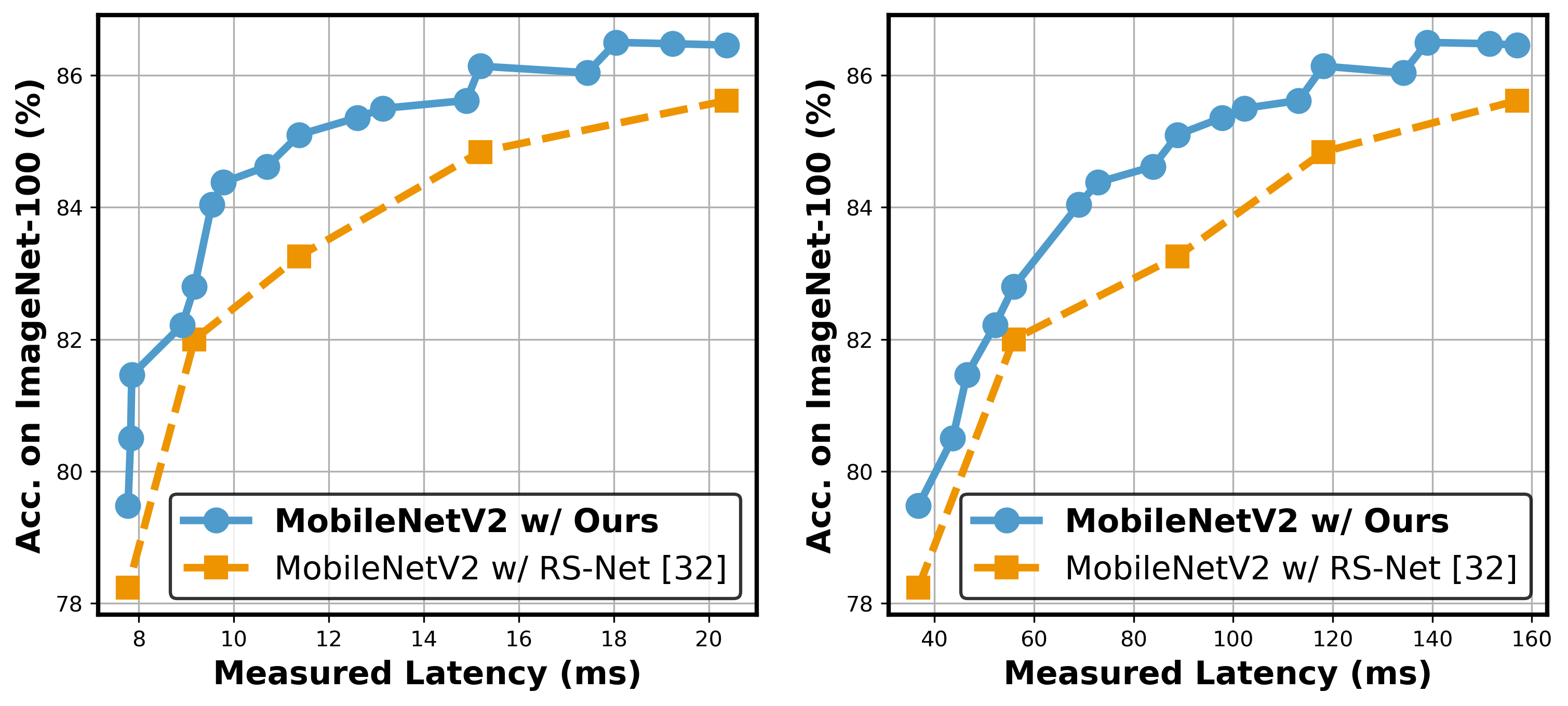}
    \end{center}
    \vspace{-4pt}
    \caption{Arbitrary-resolution elastic training vs. resolution switchable network (RS-Net) \cite{wang2020resolution} across various input resolutions on Xavier (\textit{left}) and Nano (\textit{right}) with MobileNetV2 as the baseline network.}
    \vspace{-6pt}
    \label{fig:elastic-results-against-rsnet}
\end{figure}

\textbf{Results against RS-Net \cite{wang2020resolution}.}
As shown in Fig.~\ref{fig:comparisons-overview-rsnet-ours} (\textit{left}), RS-Net features switchable batch normalization to achieve runtime resolution switching towards elastic accuracy-efficiency trade-offs. However, RS-Net can only generalize to a limited number of input resolutions since each input resolution must come with an independent batch normalization layer \cite{yu2018slimmable}. In contrast to RS-Net, the proposed elastic training technique focuses on enabling natural network elasticity across arbitrary input resolutions. Taking MobileNetV2 as an example, we compare the proposed arbitrary-resolution elastic training technique against RS-Net. The experimental results are illustrated in Fig.~\ref{fig:elastic-results-against-rsnet}, which clearly shows that the proposed arbitrary-resolution elastic training technique can consistently achieve better accuracy on ImageNet-100 across various input resolutions from $96\times96$ to $224\times224$.

\begin{figure}[t]
    \begin{center}
        \includegraphics[width=1.0\columnwidth]{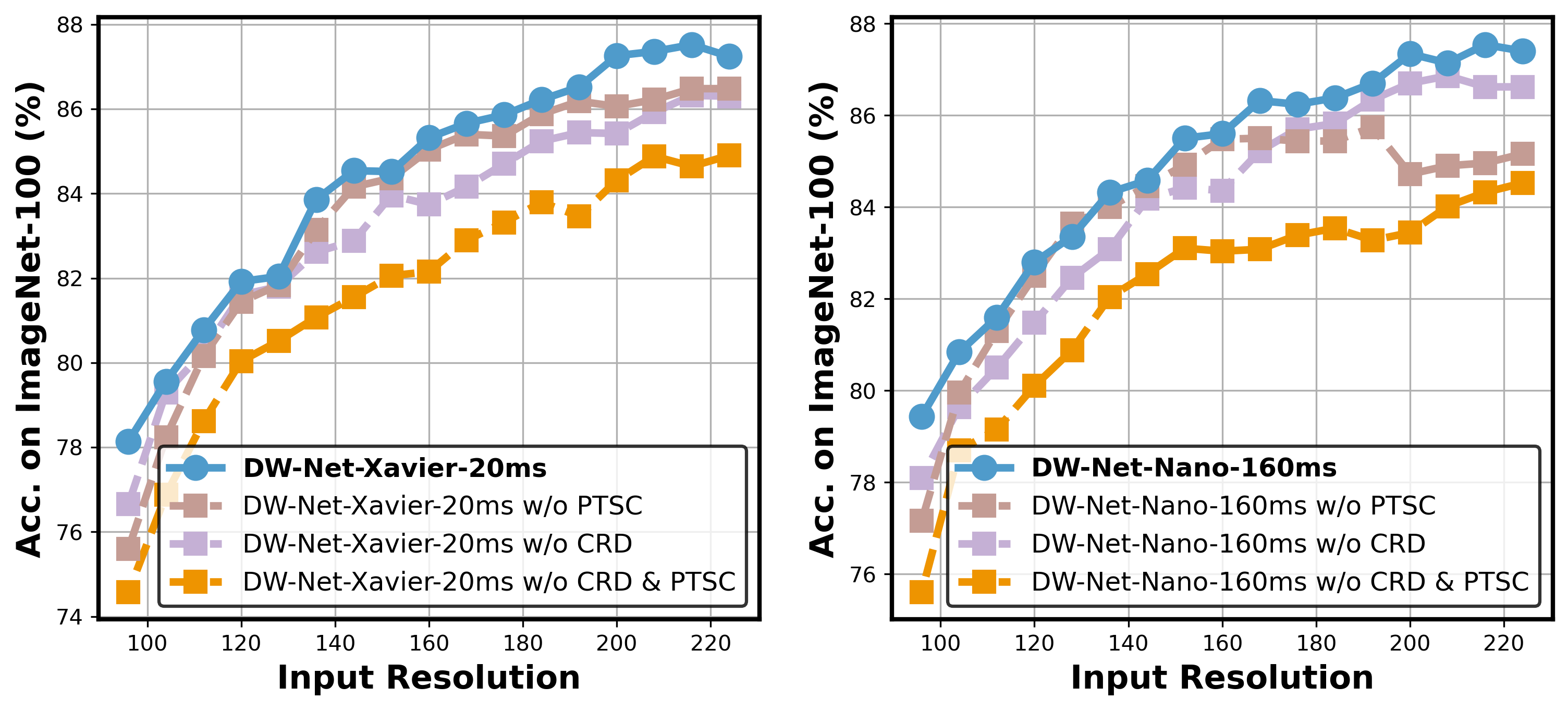}
    \end{center}
    \vspace{-4pt}
    \caption{Factorial ablation results of arbitrary-resolution training on Xavier (\textit{left}) and Nano (\textit{right}) where CRD denotes cross-resolution distillation and PTSC denotes post-training statistics calibration.}
    \vspace{-6pt}
    \label{fig:ablation-elastic-training}
\end{figure}

\textbf{Ablation of Elastic Training.}
As discussed in Section~\ref{sec:hybrid-any-resolution-elastic-training}, we introduce three components to enable dynamic accuracy-efficiency trade-offs, including (1) arbitrary-resolution elastic training, (2) cross-resolution distillation, and (3) post-training statistics calibration, where arbitrary-resolution elastic training serves as a prerequisite. Specifically, for cross-resolution distillation, we leverage the output logits from the largest input resolution to enhance the training process of subsequent small input resolutions. For post-training statistics calibration, we employ 10,000 training images to accumulate the running mean and variance statistics of batch normalization layers for each input resolution at the end of training. Note that post-training statistics calibration requires inference without computation-intensive backward propagation, which can be done offline and only takes seconds on one GeForce RTX 3090 GPU. To investigate the efficacy of each component, we further conduct a series of factorial ablation experiments. As illustrated in Fig.~\ref{fig:ablation-elastic-training}, we can observe that both cross-resolution distillation and post-training statistics calibration can largely enhance the training accuracy on ImageNet-100.

\begin{figure}[t]
    \begin{center}
        \includegraphics[width=1.0\columnwidth]{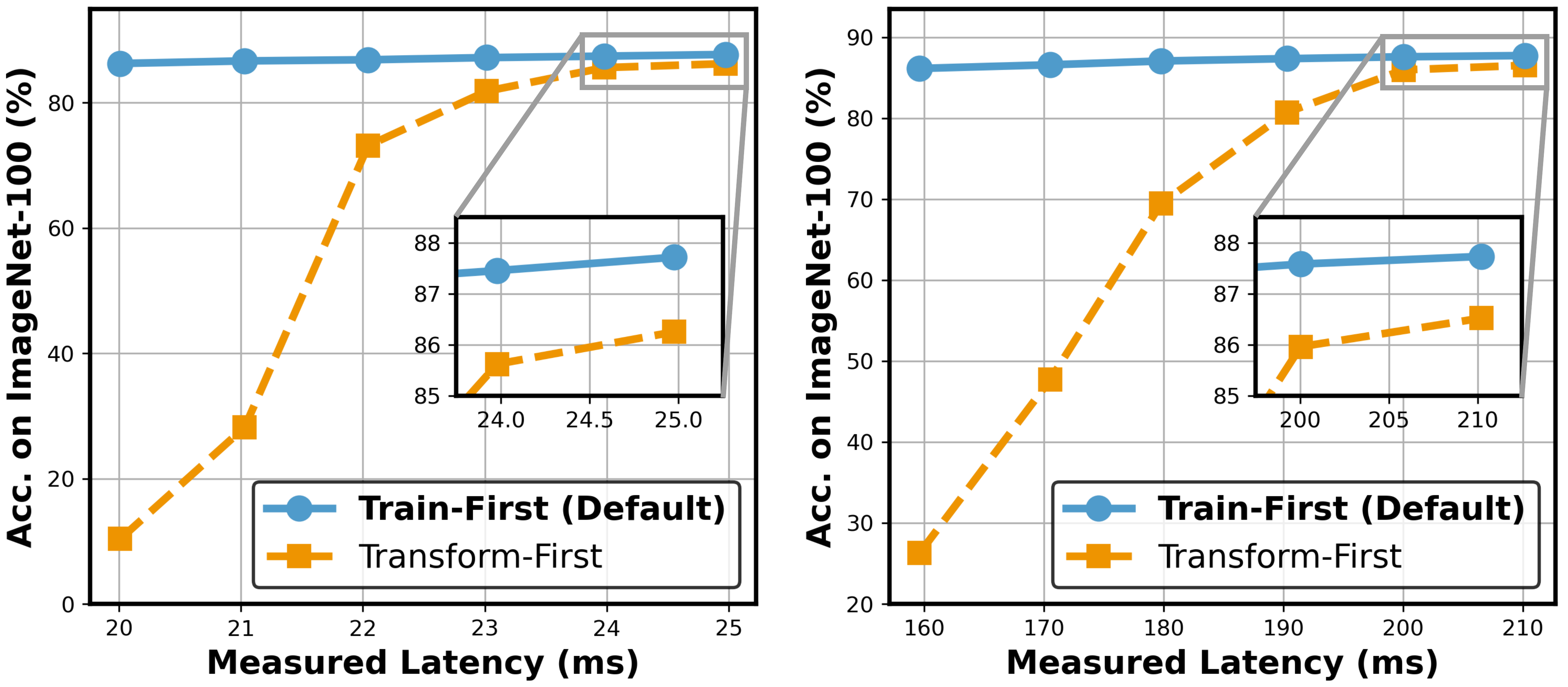}
    \end{center}
    \vspace{-4pt}
    \caption{Train-First vs. Transform-First on Xavier (\textit{left}) / Nano (\textit{right}).}
    \vspace{-6pt}
    \label{fig:train-first-vs-transform-first}
\end{figure}

\begin{figure}[t]
    \begin{center}
        \includegraphics[width=1.0\columnwidth]{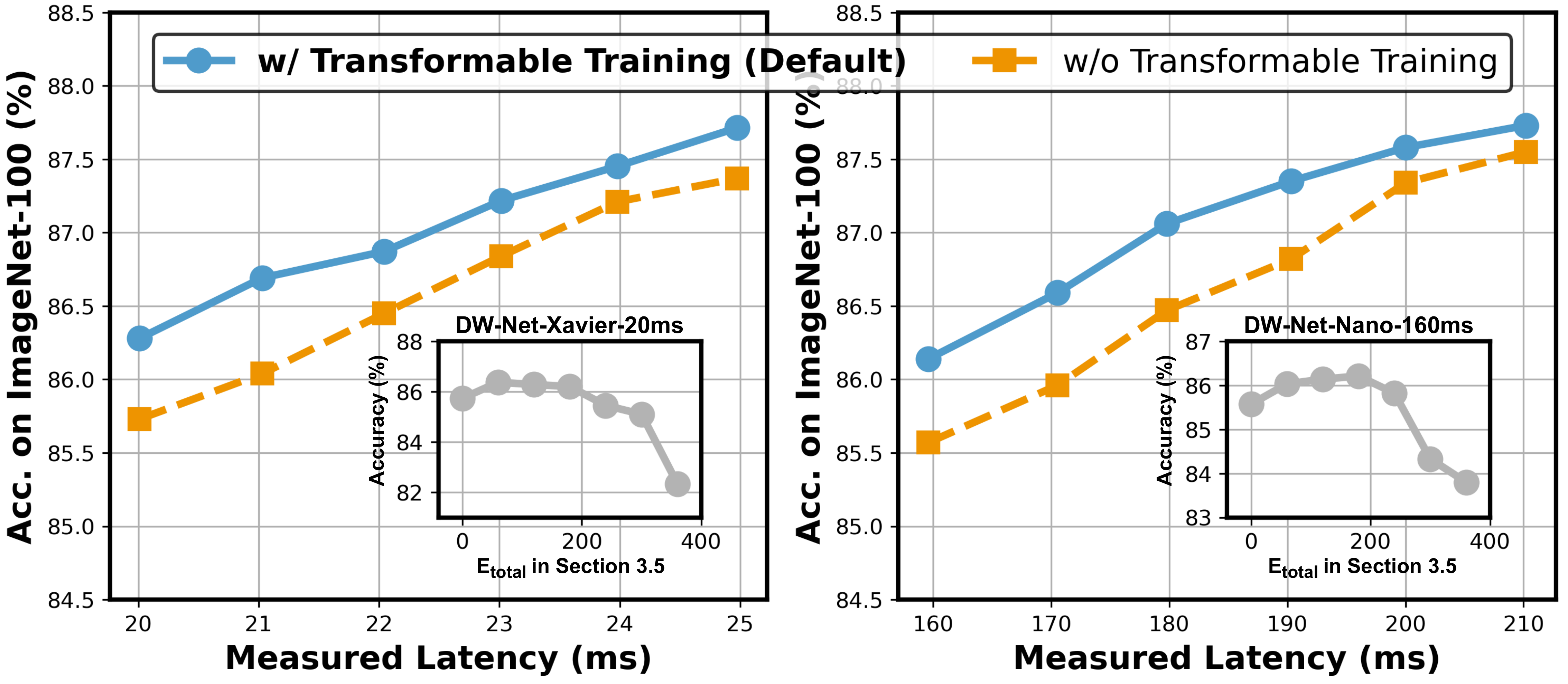}
    \end{center}
    \vspace{-4pt}
    \caption{Ablation of hybrid transformable training on Xavier (\textit{left}) and Nano (\textit{right}), where zoom-in figures show different $E_{total}$ settings.}
    \vspace{-6pt}
    \label{fig:ablation-hybrid-transformable-training}
\end{figure}

\begin{figure}[t]
    \begin{center}
        \includegraphics[width=1.0\columnwidth]{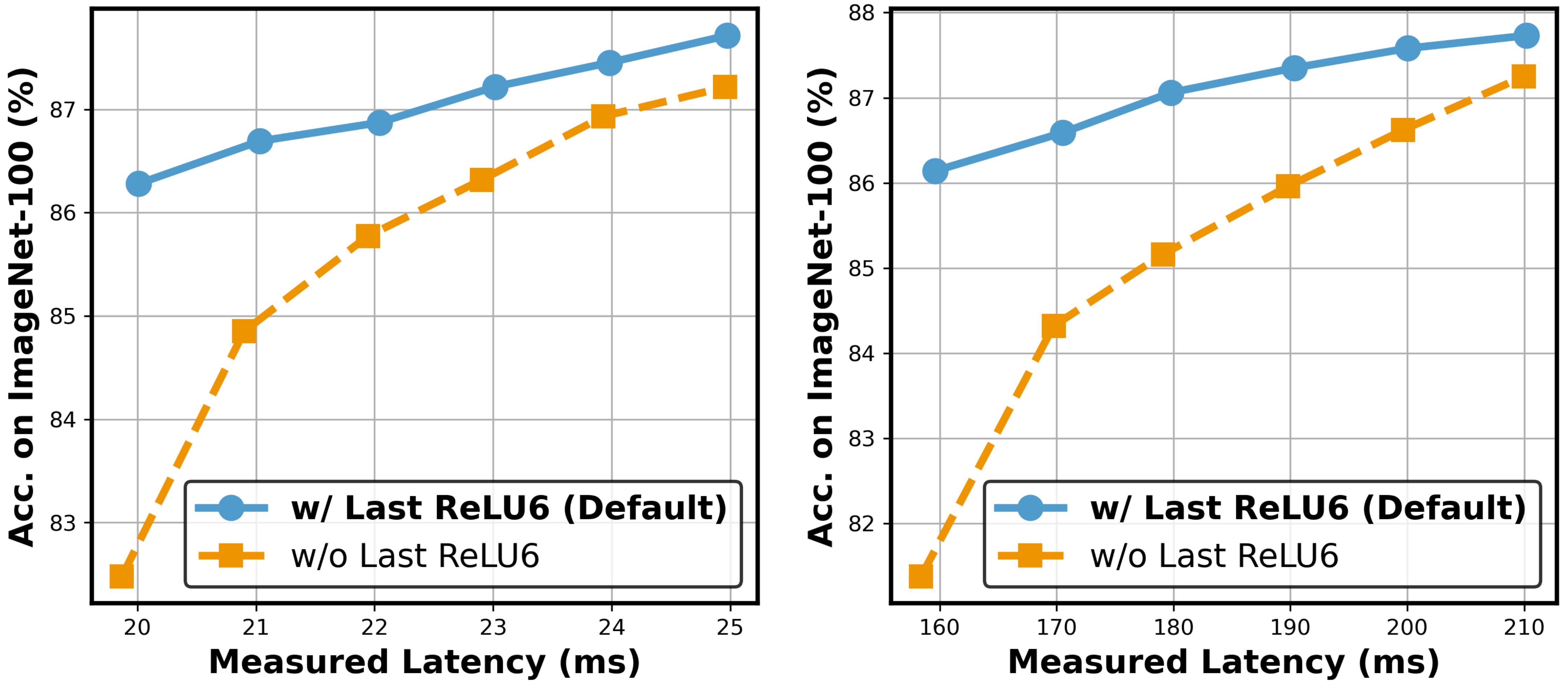}
    \end{center}
    \vspace{-4pt}
    \caption{Ablation of last ReLU6 on Xavier (\textit{left}) and Nano (\textit{right}).}
    \vspace{-6pt}
    \label{fig:ablation-last-relu6}
\end{figure}

\textbf{Train-First vs. Transform-First.}
As shown in Fig.~\ref{fig:overview-htt} (\textit{left}), we first train the searched transformable network with sufficient network depth, which is then equivalently transformed into its shallow counterpart (i.e., Train-First). In parallel, another natural training option is to first transform the searched transformable network into its shallow counterpart to eliminate its linear network redundancy and then train the resulting shallow network instead (i.e., Transform-First). The above two training options can end up with the same shallow network structure and thus maintain the same runtime efficiency on target hardware. However, as illustrated in Fig.~\ref{fig:train-first-vs-transform-first}, the latter may suffer from considerable accuracy loss on ImageNet-100, which reveals that the linear network redundancy can enhance the hybrid transformable training process towards better training accuracy.

\textbf{Ablation of Hybrid Transformable Training.}
We next investigate the efficacy of hybrid transformable training. To this end, we exclude hybrid transformable training and re-train all the searched transformable networks from scratch on ImageNet-100. Note that the searched transformable network can always be transformed into the same shallow network structure regardless of hybrid transformable training (see Fig.~\ref{fig:overview-htt}). As shown in Fig.~\ref{fig:ablation-hybrid-transformable-training}, hybrid transformable training can deliver consistent accuracy gains on ImageNet-100, while also maintaining the same on-device efficiency on both Xavier and Nano. In addition, as shown in the zoom-in figures, $E_{total} \in [60, 180]$ allows us to safely eliminate the grafted network non-linearity and convert the grafted transformable network back to the original transformable network for subsequent transformation. In light of the above experimental results, we empirically set the default $E_{total}$ to 120.

\textbf{Ablation of Last ReLU6.}
As discussed in Section~\ref{sec:hybrid-transformable-search-space}, we introduce an additional non-linear ReLU6 activation at the end of each linear \texttt{MBConv} operator, which is computationally cheap as shown in Fig.~\ref{fig:mobilenetv2-profiling-results}. It is worth noting that each linear \texttt{MBConv} operator with multiple consecutive linear layers can still be equivalently transformed into one single \texttt{Conv} layer. As illustrated in Fig.~\ref{fig:ablation-last-relu6}, this technique can largely enhance the searched transformable network towards better training accuracy on ImageNet-100, which only introduces negligible computational cost on both Xaier and Nano.

\begin{table}[t]
\centering
\caption{Ablation of the sandwich rule on ImageNet-100.}
\resizebox{1.0\linewidth}{!}{%
\begin{tabular}{l|c|c|c}
\toprule[0.125em]
\multirow{2}{*}{Differentiable Search Algorithm} &
  \multirow{2}{*}{\begin{tabular}[c]{@{}c@{}}Search Cost\\ (GPU-Days) $\downarrow$\end{tabular}} &
  \multirow{2}{*}{\begin{tabular}[c]{@{}c@{}}Latency \\ (ms) $\downarrow$\end{tabular}} &
  \multirow{2}{*}{\begin{tabular}[c]{@{}c@{}}Accuracy \\ (\%) $\uparrow$\end{tabular}} \\
                                  &     &      &      \\ \hline
Vanilla Gumbel-Softmax \cite{jang2016gumbel-softmax}                    & OOM & OOM & OOM \\ 
Single-Path Gumbel-Softmax \cite{dong2019searching}                   & 0.4 & 20.0 & 85.76 \\ 
Gumbel-Softmax Top-$k$ w/o Sandwich \cite{kool2019gumbel-softmax-topk} & 2.1 & 20.0 & 86.42 \\ \hline
\cellcolor{Gray}Gumbel-Softmax Top-$k$ w/ Sandwich (ours)  & \cellcolor{Gray}0.6 & \cellcolor{Gray}20.0 & \cellcolor{Gray}86.28 \\ 
\toprule[0.125em]
\end{tabular}%
}
\label{tab:ablation-of-sandwich}
\end{table}

\textbf{Ablation of Sandwich Rule}
Furthermore, we compare the proposed differentiable search algorithm featuring the sandwich rule with another three natural search counterparts \cite{jang2016gumbel-softmax, dong2019searching, kool2019gumbel-softmax-topk}. Among them, vanilla Gumbel-Softmax \cite{jang2016gumbel-softmax} simultaneously optimizes all the $|\mathcal{O}|=12$ operator candidates in the supernet, which inevitably involves considerable memory consumption during the search process and suffers from out-of-memory (OOM) search crashes. The experimental results are summarized in TABLE~\ref{tab:ablation-of-sandwich}, which clearly shows that the proposed sandwich rule can achieve reliable search performance while also maintaining strong search efficiency.

\begin{table}[t]
\centering
\caption{Comparisons with previous SOTA detection backbones on COCO2017, where the reported latency is measured on Xavier.}
\resizebox{1.0\linewidth}{!}{%
\begin{tabular}{l|c|c|c|c|c|c|c}
\toprule[0.125em]
Network & $\mathrm{AP}$ & $\mathrm{AP_{50}}$ & $\mathrm{AP_{75}}$ & $\mathrm{AP_S}$ & $\mathrm{AP_M}$ & $\mathrm{AP_L}$ & Latency (ms) \\ \hline
ProxylessNAS \cite{cai2018proxylessnas}       & 20.3 & 34.6 & 20.3 & 2.2 & 19.3 & 39.6 & 70.1 \\
MobileNetV2 \cite{sandler2018mobilenetv2}        & 20.4 & 34.3 & 20.5 & 1.6 & 19.5 & 40.2 & 72.6 \\
FBNet-C \cite{wu2019fbnet}              & 21.5 & 36.2 & 21.9 & 2.5 & 20.9 & 41.5 & 76.5 \\
\hline
\cellcolor{Gray}DW-Net-Xavier-20ms & \cellcolor{Gray}21.6   & \cellcolor{Gray}36.5   & \cellcolor{Gray}21.9   & \cellcolor{Gray}2.4  & \cellcolor{Gray}21.5   & \cellcolor{Gray}42.1   & \cellcolor{Gray}66.3   \\
\toprule[0.125em]
\end{tabular}%
}
\vspace{-5pt}
\label{tab:object-detection}
\end{table}

\textbf{Results on Downstream Tasks.}
We also evaluate the searched deep-to-shallow transformable networks on downstream tasks, in which we take object detection as an example. Specifically, we employ SSDLite as the default detection engine and insert different networks as drop-in backbone replacements. Note that all the networks are trained from scratch on COCO2017 using the same training recipe for fair comparisons. The experimental results are summarized in TABLE~\ref{tab:object-detection}, which clearly shows that DW-Net-Xavier-20ms can achieve better detection accuracy-efficiency trade-offs than previous SOTA detection backbones \cite{cai2018proxylessnas, wu2019fbnet, sandler2018mobilenetv2}.

\section{Conclusion}
\label{sec:conclusion}

In this paper, we present the first-of-its-kind deep-to-shallow transformable NAS paradigm, namely Double-Win NAS (DW-NAS), which provides a holistic solution for designing and training deep-to-shallow transformable networks, marrying the best of both deep and shallow networks towards an aggressive win-win in terms of both accuracy and hardware efficiency. Specifically, DW-NAS strives to explore deep networks with sufficient network depth to first win superior accuracy, which then can be equivalently transformed into their shallow counterparts to further win considerable hardware efficiency without accuracy loss. Furthermore, we also introduce two enhanced training techniques, including hybrid transformable training towards better training accuracy and hybrid arbitrary-resolution training towards runtime resolution switching across arbitrary input resolutions. Extensive experiments on two popular NVIDIA Jetson intelligent embedded systems and two representative large-scale datasets clearly demonstrate the superiority of DW-NAS over previous SOTA NAS approaches.

\balance
\bibliographystyle{unsrt}
\bibliography{reference}

@article{liu2022bringing,
  title={{Bringing AI To Edge: From Deep Learning's Perspective}},
  author={Liu et al., Di},
  journal={Neurocomputing},
  year={2022},
}

@article{luo2024efficient,
  title={{Efficient Deep Learning Infrastructures for Embedded Computing Systems: A Comprehensive Survey and Future Envision}},
  author={Luo et al., Xiangzhong},
  journal={ACM Transactions on Embedded Computing Systems (TECS)},
  year={2024},
}

@article{luo2024domino,
  title={{Domino-Pro-Max: Toward Efficient Network Simplification and Reparameterization for Embedded Hardware Systems}},
  author={Luo et al., Xiangzhong},
  journal={IEEE Transactions on Computer-Aided Design of Integrated Circuits and Systems (TCAD)},
  year={2024},
}

@inproceedings{li2022rt,
  title={{RT-NeRF: Real-Time On-Device Neural Radiance Fields Towards Immersive AR/VR Rendering}},
  author={Li et al., Chaojian},
  booktitle={IEEE/ACM International Conference on Computer-Aided Design (ICCAD)},
  year={2022}
}

@inproceedings{neseem2021adacon,
  title={{AdaCon: Adaptive Context-Aware Object Detection for Resource-Constrained Embedded Devices}},
  author={Neseem et al., Marina},
  booktitle={IEEE/ACM International Conference On Computer Aided Design (ICCAD)},
  year={2021},
}

@inproceedings{he2016deep,
  title={{Deep Residual Learning for Image Recognition}},
  author={He et al., Kaiming},
  booktitle={Proceedings of the IEEE Conference on Computer Vision and Pattern Recognition (CVPR)},
  year={2016}
}

@inproceedings{he2015convolutional,
  title={{{Convolutional Neural Network at Constrained Time Cost}}},
  author={He et al., Kaiming},
  booktitle={Proceedings of the IEEE Conference on Computer Vision and Pattern Recognition (CVPR)},
  year={2015}
}

@article{goyal2022non,
  title={{Non-Deep Networks}},
  author={Goyal et al., Ankit},
  journal={Advances in Neural Information Processing Systems (NeurIPS)},
  year={2022}
}

@inproceedings{sandler2018mobilenetv2,
  title={{MobileNetV2: Inverted Residuals and Linear Bottlenecks}},
  author={Sandler et al., Mark},
  booktitle={Proceedings of the IEEE Conference on Computer Vision and Pattern Recognition (CVPR)},
  year={2018}
}

@article{liu2018darts,
  title={{DARTS: Differentiable Architecture Search}},
  author={Liu et al., Hanxiao},
  journal={International Conference on Learning Representations (ICLR)},
  year={2019}
}

@inproceedings{tan2019mnasnet,
  title={{MnasNet: Platform-Aware Neural Architecture Search for Mobile}},
  author={Tan et al., Mingxing},
  booktitle={Proceedings of the IEEE/CVF Conference on Computer Vision and Pattern Recognition (CVPR)},
  year={2019}
}

@inproceedings{chu2020moga,
  title={{MoGA: Searching Beyond MobileNetV3}},
  author={Chu et al., Xiangxiang},
  booktitle={IEEE International Conference on Acoustics, Speech and Signal Processing (ICASSP)},
  year={2020},
}

@inproceedings{howard2019searching,
  title={{Searching for MobileNetV3}},
  author={Howard et al., Andrew},
  booktitle={Proceedings of the IEEE/CVF International Conference on Computer Vision (ICCV)},
  year={2019}
}

@article{cai2019once,
  title={{Once-for-All: Train One Network and Specialize It for Efficient Deployment}},
  author={Cai et al., Han},
  journal={International Conference on Learning Representations (ICLR)},
  year={2020}
}

@inproceedings{ma2018shufflenet,
  title={{ShuffleNetV2: Practical Guidelines for Efficient CNN Architecture Design}},
  author={Ma et al., Ningning},
  booktitle={Proceedings of the European Conference on Computer Vision (ECCV)},
  year={2018}
}

@inproceedings{zhang2021dian,
  title={{DIAN: Differentiable Accelerator-Network Co-Search Towards Maximal DNN Efficiency}},
  author={Zhang et al., Yongan},
  booktitle={IEEE/ACM International Symposium on Low Power Electronics and Design (ISLPED)},
  year={2021},
}

@inproceedings{hu2020tfnas,
  title={{TF-NAS: Rethinking Three Search Freedoms of Latency-Constrained Differentiable Neural Architecture Search}},
  author={Hu et al., Yibo},
  booktitle={Proceedings of the European Conference on Computer Vision (ECCV)},
  year={2020},
}

@inproceedings{wu2019fbnet,
  title={{FBNet: Hardware-Aware Efficient ConvNet Design via Differentiable Neural Architecture Search}},
  author={Wu et al., Bichen},
  booktitle={Proceedings of the IEEE/CVF Conference on Computer Vision and Pattern Recognition (CVPR)},
  year={2019}
}

@article{cai2018proxylessnas,
  title={{ProxylessNAS: Direct Neural Architecture Search on Target Task and Hardware}},
  author={Cai et al., Han},
  journal={International Conference on Learning Representations (ICLR)},
  year={2019}
}

@inproceedings{stamoulis2019single,
  title={{Single-Path NAS: Designing Hardware-Efficient ConvNets in Less Than 4 Hours}},
  author={Stamoulis et al., Dimitrios},
  booktitle={Joint European Conference on Machine Learning and Knowledge Discovery in Databases (ECML-PKDD)},
  year={2019},
}

@inproceedings{fang2020densely,
  title={{Densely Connected Search Space for More Flexible Neural Architecture Search}},
  author={Fang et al., Jiemin},
  booktitle={Proceedings of the IEEE/CVF Conference on Computer Vision and Pattern Recognition (CVPR)},
  year={2020}
}

@inproceedings{li2020edd,
  title={{EDD: Efficient Differentiable DNN Architecture and Implementation Co-search for Embedded AI Solutions}},
  author={Li et al., Yuhong},
  booktitle={ACM/IEEE Design Automation Conference (DAC)},
  year={2020},
}

@inproceedings{luo2022you,
  title={{You Only Search Once: On Lightweight Differentiable Architecture Search for Resource-Constrained Embedded Platforms}},
  author={Luo et al., Xiangzhong},
  booktitle={ACM/IEEE Design Automation Conference (DAC)},
  year={2022}
}

@article{luo2022lightnas,
  title={{LightNAS: On Lightweight and Scalable Neural Architecture Search for Embedded Platforms}},
  author={Luo et al., Xiangzhong},
  journal={IEEE Transactions on Computer-Aided Design of Integrated Circuits and Systems (TCAD)},
  year={2022},
}

@inproceedings{dong2019searching,
  title={{Searching for A Robust Neural Architecture in Four GPU Hours}},
  author={Dong et al., Xuanyi},
  booktitle={Proceedings of the IEEE/CVF Conference on Computer Vision and Pattern Recognition (CVPR)},
  year={2019}
}

@article{benmeziane2021comprehensive,
  title={{A Comprehensive Survey on Hardware-Aware Neural Architecture Search}},
  author={Benmeziane et al., Hadjer},
  journal={arXiv preprint arXiv:2101.09336},
  year={2021}
}

@inproceedings{tan2019efficientnet,
  title={{EfficientNet: Rethinking Model Scaling for Convolutional Neural Networks}},
  author={Tan et al., Mingxing},
  booktitle={International Conference on Machine Learning (ICML)},
  year={2019},
}

@inproceedings{chu2021fairnas,
  title={{FairNAS: Rethinking Evaluation Fairness of Weight Sharing Neural Architecture Search}},
  author={Chu et al., Xiangxiang},
  booktitle={Proceedings of the IEEE/CVF International Conference on Computer Vision (ICCV)},
  year={2021}
}

@inproceedings{kool2019gumbel-softmax-topk,
  title={{Stochastic Beams and Where to Find Them: The Gumbel-Top-k Trick for Sampling Sequences Without Replacement}},
  author={Kool et al., Wouter},
  booktitle={International Conference on Machine Learning (ICML)},
  year={2019},
}

@article{jang2016gumbel-softmax,
  title={{Categorical Reparameterization with Gumbel-Softmax}},
  author={Jang et al., Eric},
  journal={International Conference on Learning Representations (ICLR)},
  year={2016}
}

@article{bengio2013estimating,
  title={{Estimating or Propagating Gradients Through Stochastic Neurons for Conditional Computation}},
  author={Bengio et al., Yoshua},
  journal={arXiv preprint arXiv:1308.3432},
  year={2013}
}

@inproceedings{li2022physics,
  title={{Physics-Aware Differentiable Discrete Codesign for Diffractive Optical Neural Networks}},
  author={Li et al., Yingjie},
  booktitle={IEEE/ACM International Conference on Computer-Aided Design (ICCAD)},
  year={2022}
}

@inproceedings{hu2018squeeze,
  title={{Squeeze-and-Excitation Networks}},
  author={Hu et al., Jie},
  booktitle={Proceedings of the IEEE Conference on Computer Vision and Pattern Recognition (CVPR)},
  year={2018}
}

@article{guo2020expandnets,
  title={{ExpandNets: Linear Over-Parameterization to Train Compact Convolutional Networks}},
  author={Guo et al., Shuxuan},
  journal={Advances in Neural Information Processing Systems (NeurIPS)},
  year={2020}
}

@article{chen2023vanillanet,
  title={{VanillaNet: the Power of Minimalism in Deep Learning}},
  author={Chen et al., Hanting},
  journal={Advances in Neural Information Processing Systems (NeurIPS)},
  year={2023}
}

@inproceedings{luo2024pearls,
  title={{Pearls Hide Behind Linearity: Simplifying Deep Convolutional Networks for Embedded Hardware Systems via Linearity Grafting}},
  author={Luo et al., Xiangzhong},
  booktitle={IEEE Asia and South Pacific Design Automation Conference (ASP-DAC)},
  year={2024},
}

@misc{tensorrt,
  title = {{NVIDIA TensorRT}},
  howpublished = {\url{https://developer.nvidia.com/tensorrt}},
}

@article{dong2020bench,
  title={{NAS-Bench-201: Extending the Scope of Reproducible Neural Architecture Search}},
  author={Dong et al., Xuanyi},
  journal={International Conference on Learning Representations (ICLR)},
  year={2020}
}

@article{li2021hw,
  title={{HW-NAS-Bench: Hardware-Aware Neural Architecture Search Benchmark}},
  author={Li et al., Chaojian},
  journal={International Conference on Learning Representations (ICLR)},
  year={2021}
}

@inproceedings{wang2020resolution,
  title={{Resolution Switchable Networks for Runtime Efficient Image Recognition}},
  author={Wang et al., Yikai},
  booktitle={Proceedings of the European Conference on Computer Vision (ECCV)},
  year={2020},
}

@article{yu2018slimmable,
  title={{Slimmable Neural Networks}},
  author={Yu et al., Jiahui},
  journal={Internaltional Conference on Learning Representations (ICLR)},
  year={2018}
}

@inproceedings{yu2019universally,
  title={{Universally Slimmable Networks and Improved Training Techniques}},
  author={Yu et al., Jiahui},
  booktitle={Proceedings of the IEEE/CVF International Conference on Computer Vision (ICCV)},
  year={2019}
}

@inproceedings{wang2018skipnet,
  title={{SkipNet: Learning Dynamic Routing in Convolutional Networks}},
  author={Wang et al., Xin},
  booktitle={Proceedings of the European Conference on Computer Vision (ECCV)},
  year={2018}
}

@inproceedings{yang2022once,
  title={{Once For All Skip: Efficient Adaptive Deep Neural Networks}},
  author={Yang et al., Yu},
  booktitle={Design, Automation \& Test in Europe Conference \& Exhibition (DATE)},
  year={2022},
}

@article{touvron2019fixing,
  title={{Fixing the Train-Test Resolution Discrepancy}},
  author={Touvron et al., Hugo},
  journal={Advances in Neural Information Processing Systems (NeurIPS)},
  year={2019}
}

@article{hoffer2019mix,
  title={{Mix \& Match: Training ConvNets with Mixed Image Sizes for Improved Accuracy, Speed and Scale Resiliency}},
  author={Hoffer et al., Elad},
  journal={arXiv preprint arXiv:1908.08986},
  year={2019}
}

@article{jiang2021eh,
  title={{EH-DNAS: End-to-End Hardware-aware Differentiable Neural Architecture Search}},
  author={Jiang et al., Qian},
  journal={International Conference on Machine Learning (ICML) Workshop},
  year={2023}
}

@article{hu2024l,
  title={{l-DARTS: Lightweight Differentiable Architecture Search with Robustness Enhancement Strategy}},
  author={Hu et al., Liwei},
  journal={Knowledge-Based Systems (KBS)},
  year={2024},
}

@article{lee2024d2nas,
  title={{D2NAS: Efficient Neural Architecture Search With Performance Improvement and Model Size Reduction for Diverse Tasks}},
  author={Lee et al., Jungeun},
  journal={IEEE Access},
  year={2024},
}

@inproceedings{yang2021netadaptv2,
  title={{NetAdaptV2: Efficient Neural Architecture Search with Fast Super-Network Training and Architecture Optimization}},
  author={Yang et al., Tien-Ju},
  booktitle={Proceedings of the IEEE/CVF Conference on Computer Vision and Pattern Recognition (CVPR)},
  year={2021}
}

@article{nguyen2020wide,
  title={{Do Wide and Deep Networks Learn the Same Things? Uncovering How Neural Network Representations Vary with Width and Depth}},
  author={Nguyen et al., Thao},
  journal={arXiv preprint arXiv:2010.15327},
  year={2020}
}

@inproceedings{xue2022go,
  title={{Go Wider Instead of Deeper}},
  author={Xue et al., Fuzhao},
  booktitle={Proceedings of the AAAI Conference on Artificial Intelligence (AAAI)},
  year={2022}
}

@inproceedings{ghebriout2024harmonic,
  title={{Harmonic-NAS: Hardware-Aware Multimodal Neural Architecture Search on Resource-Constrained Devices}},
  author={Ghebriout et al., Mohamed Imed Eddine},
  booktitle={Asian Conference on Machine Learning (ACML)},
  year={2024},
}

@article{jiang2025score,
  title={{Score Predictor-Assisted Evolutionary Neural Architecture Search}},
  author={Jiang et al., Pengcheng},
  journal={IEEE Transactions on Emerging Topics in Computational Intelligence (TETCI)},
  year={2025}
}

@article{nasir2025esm,
  title={{ESM: A Framework for Building Effective Surrogate Models for Hardware-Aware Neural Architecture Search}},
  author={Nasir et al., Azaz-Ur-Rehman},
  journal={arXiv preprint arXiv:2508.01505},
  year={2025}
}

@article{fayyazi2025marco,
  title={{MARCO: Hardware-Aware Neural Architecture Search for Edge Devices with Multi-Agent Reinforcement Learning and Conformal Prediction Filtering}},
  author={Fayyazi et al., Arya},
  journal={arXiv preprint arXiv:2506.13755},
  year={2025}
}

@article{lv2025situ,
  title={{In-situ NAS: A Plug-and-Search Neural Architecture Search framework across hardware platforms}},
  author={Lv et al., Hao},
  journal={IEEE Transactions on Computers},
  year={2025},
}

\end{document}